%% file: main-ieee.tex
\documentclass[lettersize, journal]{IEEEtran}
\usepackage{amsmath,amsfonts}
\usepackage{array}
\usepackage{textcomp}
\usepackage{stfloats}
\usepackage{url}
\usepackage{verbatim}
\usepackage{graphicx}
\usepackage{balance}
\usepackage{xcolor}

\usepackage{bbding}
\usepackage{wrapfig}
\usepackage{colortbl}

\usepackage{makecell}
\usepackage{algorithm}
\usepackage{algpseudocode}
\usepackage{enumitem}
\usepackage{subcaption}

\newcommand{\name}{{\tt pNav}}

\begin{document}
\title{Power-Efficient Autonomous Mobile Robots}
\author{
\IEEEauthorblockN{
Liangkai~Liu\IEEEauthorrefmark{1},
Weisong~Shi\IEEEauthorrefmark{2}, and
Kang~G.~Shin\IEEEauthorrefmark{1}
}
\\
\IEEEauthorblockA{
\IEEEauthorrefmark{1}Department of Computer Science and Engineering, University of Michigan, USA
\\
\IEEEauthorrefmark{2}Department of Computer and Information Sciences, University of Delaware, USA
}
}


\maketitle

\begin{abstract}
\input{contents/0_Abstract}
\end{abstract}

\begin{IEEEkeywords}
power efficiency, autonomous mobile robotics, cyber-physical systems
\end{IEEEkeywords}

\section{Introduction}
\input{contents/1_Introduction}

\section{Background and Motivation}
\input{contents/2_Background_Motivation}

\section{Challenges in Achieving CPS Power-Efficiency}
\input{contents/3_Empirical_Studies}

\section{Design of \texttt{pNav}}
\input{contents/4_System_Design}

\section{Design of Donkey, a Power-Profiling Platform}
\input{contents/Donkey}

\section{Implementation}
\input{contents/5_Implementation}

\section{Evaluation}
\input{contents/6_Evaluation}


\vspace{-3mm}
\section{Related Work}
\input{contents/8_Related_Works}

\section{Conclusion}
\input{contents/9_Conclusion}


\bibliographystyle{IEEEtran}
\bibliography{main}

\end{document}

%% file: contents/0_Abstract.tex
This paper presents \name, a novel power-management system that significantly 
enhances the power/energy-efficiency of Autonomous Mobile Robots (AMRs) 
by jointly optimizing their physical/mechanical and cyber subsystems. 
By profiling AMRs' power consumption, we identify three challenges in 
achieving CPS (cyber-physical system) power-efficiency that involve 
both cyber (C) and physical (P) subsystems: 
(1) variabilities of system power consumption breakdown, 
(2) environment-aware navigation locality, and (3) coordination 
of C and P subsystems. 
\name\ takes a multi-faceted approach to achieve power-efficiency of AMRs.
First, it integrates millisecond-level power consumption prediction for 
both C and P subsystems. Second, it includes novel real-time modeling 
and monitoring of spatial and temporal navigation localities for AMRs. 
Third, it supports dynamic coordination of AMR software 
(navigation, detection) and hardware (motors, DVFS driver) configurations. 
\name\ is prototyped using the Robot Operating System (ROS) 
Navigation Stack, 2D LiDAR, and camera. Our in-depth evaluation with a 
real robot and Gazebo environments demonstrates a $>$96\% accuracy in 
predicting power consumption and a 38.1\% reduction in power consumption
without compromising navigation accuracy and safety.


%% file: contents/1_Introduction.tex
\label{sec:introduction}

Autonomous mobile robots (AMRs) have been omnipresent, 
including factories for manufacturing goods and hospitals 
for food/drug delivery, personal assistance, search \& 
rescue, deep space exploration, etc.~%
\cite{siegwart2011introduction, recog, AMR_farm, AMR_DC}. 
As a prototypical cyber-physical system (CPS), 
an AMR is composed of the physical (P) subsystem
(sensors, motors, batteries, etc.) and the 
cyber (C) subsystem (computing and communication HW/SW, 
navigation algorithm/software)~\cite{lee2008cyber}. 
One of the key challenges for AMRs is their 
power-efficiency. The operation of an AMR is 
constrained by its limited battery capacity, making 
its power-efficiency critically important.

The state-of-the-art (SOTA) research on enhancing 
AMR power-efficiency has taken two main 
avenues~\cite{swanborn2020energy,farooq2023power}. 
The first involves energy-efficient motion planning, 
optimizing the power usage by physical components, 
and modeling the relationship between motor speed and 
power consumption \cite{barili1995energy, mei2005case, 
path_planning1, path_planning3, path_planning4, 
movingsensor, movingsensor2, movingsensor3, 
henkel2016energy, brateman2006energy}. However, the 
increasing use of GPUs in cyber/embedded systems makes 
it challenging to meet growing power demands 
\cite{jetson-comparison}. The second centers 
around creating energy-efficient cyber/embedded systems,
employing Dynamic Voltage and Frequency Scaling 
(DVFS) \cite{chen2015smartphone, lin2023workload, 
choi2019optimizing, bateni2020neuos, choi2019graphics}. 
However, these strategies are better suited for mobile 
electronic devices and may not meet the specific timing 
constraints of AMRs. Approaches like E2M \cite{liu2019e2m} 
aim to optimize computing power in AMRs but often overlook 
the power consumption by the mechanical/physical 
subsystem. Another proposal is an end-to-end (e2e) 
energy model for AMRs \cite{liu2023open}, but its 
simple path planning falls short and lacks full 
system integration. Thus, the central challenge in 
developing AMRs lies in the optimization of power 
consumption for {\em both} cyber (C) and the 
physical (P) subsystems. To the best of our knowledge,
this paper is the first to cover both in-depth.

\begin{figure}[t]
	\centering
	\includegraphics[trim=0cm 0cm 0cm 0cm, clip, width=.8\columnwidth]{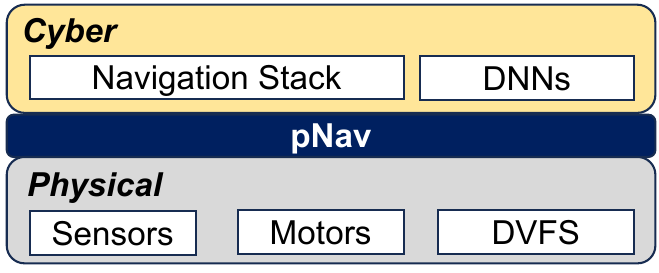}
	\caption{\name --- holistic power-management system}
	\label{fig:general-idea}
\end{figure}

Joint optimization of C and P subsystems' power consumption 
requires comprehensive power profiling of AMRs.   
Therefore, we first developed a representative robot called Donkey and a 
simulation environment for a detailed analysis of AMR's total power 
consumption. We then identified three technical challenges
in achieving power-efficiency for AMRs. 
First, there is a lack of understanding of power-breakdown 
variabilities under different operation speeds and modes. 
An efficient power-management system needs to model real-time power 
consumption of both C and P subsystems. 
Second, ignoring environment-aware navigation 
localities leads to power wastage. AMRs exhibit significant spatial 
and temporal localities, i.e., their positions and sensing data are 
similar over short periods and small spatial areas. 
Third, uncoordinated C and P subsystems lead to power wastage. 
Information from the P subsystem, such as the AMR's speed, can help 
reduce the C subsystem's power consumption, and vice versa. 
An optimal solution requires the proper 
co-design of C and P subsystems. 



We tackle the above-identified challenges by developing 
\name, a holistic power-management system that 
optimizes the power-efficiency and navigation-safety 
of AMRs with dynamic and real-time cooridnation of 
C and P subsystems. 
as shown in Fig.~\ref{fig:general-idea}. 
\name\ comprises four key components: Power Predictor, Locality Checker, 
Collision Predictor, and Coordinator. 
The Power Predictor (\S\ref{subsec:power-predictor}) estimates the 
overall power consumption of an AMR at millisecond-level 
based on motor speed, CPU/GPU frequencies, and navigation 
configurations. The prediction model is integrated 
into the AMR's path planning for navigation.
The Locality Checker (\S\ref{subsec:locality-checker}) focuses on 
modeling spatial and temporal navigation localities at three level: 
data level (robot's Field of View (FOV) overlap), postion level 
(SLAM particle confidence), and path level (trajectory 
waypoints). Utilizing LiDAR data, the Collision Predictor 
(\S\ref{subsec:collision-predictor}) 
forecasts potential collisions through timeline analysis 
for the worst-case scenario. It continually assesses 
the dynamic environment to define a safe time interval, 
enabling safe adjustments to the navigation configurations of the AMR. 
Central to \name\ is the Coordinator (\S\ref{subsec:coordinator}), which 
integrates the results from various modules to decide whether to run 
the AMR in the power-saving or performance mode. In power-saving mode, 
the Coordinator dynamically adjusts motor speed, navigation 
configurations, and DVFS settings to achieve optimal power-efficiency 
while considering the updated navigation locality and 
TTC (Time-To-Collision) constraints. When the navigation locality 
or TTC falls below a certain threshold, the Coordinator switches 
to the performance mode to enhance navigation performance. 
Once the locality and TTC values get higher, the system returns to 
the power-saving mode. \name\ effectively balances 
power-efficiency and navigation safety for AMRs.


We have prototyped \name\ with general/common AMRs hardware 
(2D LiDAR, camera) and software (ROS navigation stack, detector) 
designs~\cite{gatesichapakorn2019ros,pimentel2021evaluation,henkel2016energy,turtlebot-github,ros-navigation}. 
Then, we have conducted extensive experiments to assess \name's effectiveness
in terms of the prediction and reduction of power consumption
as well as the efficiency of AMR navigation. Our evaluation 
has shown \name\ to achieve an accuracy of over 96\% 
in e2e power-consumption prediction for AMRs. 
It also reduces power consumption by 38.1\% over the
baseline AMR configurations. 
The increase in task execution latency for \name\ remains 
under 6\%, making a minimal impact on operational speed. 
Moreover, the accuracy drop in localization was $<$2\%, 
showcasing its high navigation precision. \name\ also 
improves AMR's safety by extending the time-to-collision 
(TTC), thus letting AMRs buy time to react to and avoid 
potential hazards. Overall, \name\ significantly enhances 
the power-efficiency and navigational safety of AMRs.

This paper makes the following three main contributions:
\begin{itemize}[noitemsep,topsep=0pt,parsep=0pt,partopsep=0pt]
    \item A comprehensive and fine-grained power profiling of AMRs, 
    revealing three types of challenges in achieving power-efficient 
    AMRs system: (1) variabilities of system power consumption breakdown, 
    (2) environment-aware navigation locality, and (3) coordination 
    of C and P subsystems (\S\ref{sec:empirical-study});
    \item Development of \name, addressing the above inefficiencies. 
    \name\ combines a real-time prediction of power consumption for both 
    C and P subsystems, novel modeling and monitoring of 
    navigation localities, a collision prediction module, and 
    effective coordination of both software and hardware configurations 
    across both C and P subsystems to realize energy-efficient and 
    safe AMRs (\S\ref{sec:system});
    \item Prototyping \name\ using the ROS Navigation Stack (\S\ref{sec:implementation}), and evaluating its effectiveness
    via extensive experiments. Our results demonstrate that 
    \name\ achieves a 96\% accuracy in power-consumption 
    prediction and reduces power consumption by 38.1\% for 
    AMRs. \name\ also ensures efficient and safe navigation 
    with a minimal ($<6$\%) increase of task latency and
    a $<$2\% drop in localization accuracy (\S\ref{sec:evaluation}).
\end{itemize}

%% file: contents/2_Background_Motivation.tex
\label{sec:motivation}

\subsection{Autonomous Mobile Robots}

\begin{figure}[!htp]
	\centering
	\includegraphics[trim=0cm 0cm 0cm 0cm, clip, width=\columnwidth]{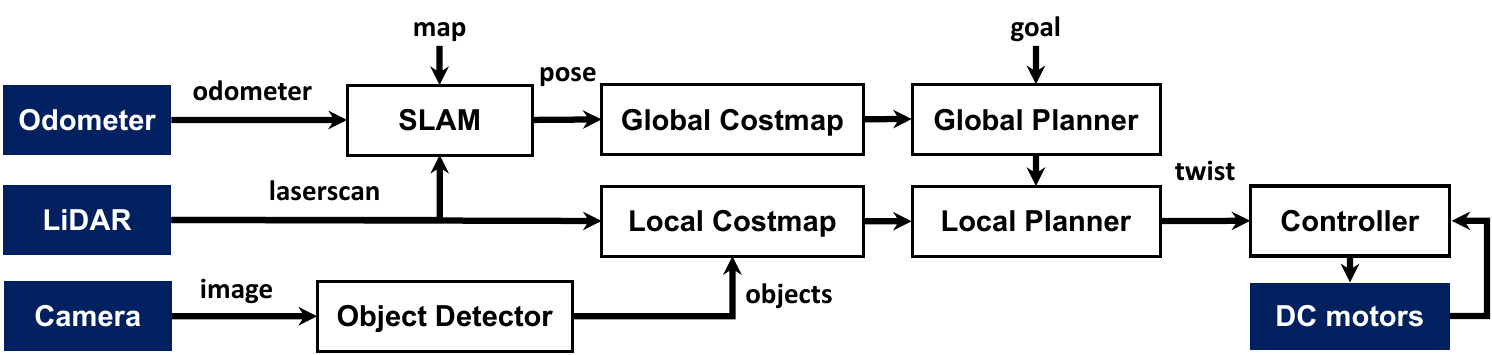}
	\caption{A general software pipeline for AMRs.}
	\label{fig:AMR-software-pipeline}
\end{figure}

\noindent\textbf{Software Pipeline:} 
Fig.~\ref{fig:AMR-software-pipeline} depicts a general software 
pipeline for AMRs~\cite{ros-navigation, turtlebot-github}.  
The pipeline starts with sensory inputs from odometry, 
LiDARs, cameras, etc., to such subsystems as Simultaneous 
Localization and Mapping (SLAM) for mapping and localization, 
and an object detector for recognizing obstacles. 
This sensory input data is processed to create both global and 
local cost-maps, which are essential for navigation. 
The core of the pipeline involves global and 
local planners, which work in tandem to build a route 
toward a designated goal while accounting for real-time 
environmental changes. Finally, the controller translates 
these plans into actionable commands for the robot's 
actuators --- typically DC motors --- to perform precise 
movement and interaction with the environment. 

\vspace{-3mm}
\subsection{Motivation}


The increasing deployment of AMRs  
in diverse sectors brings to the forefront the critical 
need for efficient power management. State-of-the-art
(SOTA) approaches to power optimization in AMRs tend to 
be siloed, focusing on either the planning algorithms or 
the cyber/embedded systems' efficiency, without a comprehensive, 
system-wide perspective~\cite{path_planning1, path_planning3, path_planning4, movingsensor, movingsensor2, movingsensor3, liu2019e2m, liu2023open}. 
This gap calls for an innovative approach that considers the 
{\em total} power consumption of each AMR, including sensors, 
motors, and computing platform, to enhance 
its endurance and operational effectiveness.
 
\textit{We aim to bridge 
this gap by developing a comprehensive solution that 
optimizes power use across all components of AMRs.} 
To the best of our knowledge, this is the first to 
holistically optimize the power consumption of both 
the C and P components of AMRs.

%% file: contents/3_Empirical_Studies.tex
\label{sec:empirical-study}

To understand and identify the technical challenges of power management 
for AMRs, we have conducted comprehensive empirical studies using Donkey 
and Turtlebot. 
We have måde three key observations: 1) variabilities of system 
power consumption breakdown, 2) environment-dependent navigation localities, 
3) coordination of cyber (C) and physical (P) subsystems to achieve 
optimal power-efficiency.

\subsection{Variabilities of System Power Consumption Breakdown}


We first break down the power consumption of the end-to-end 
(e2e) system on a real robot (Section~\ref{sec:evaluation}). 
The power-consumption breakdown varies greatly with the motor 
speed and tasks running on the cyber/embedded subsystem.
We choose the four scenarios in 
Table~\ref{tab:power-dissipation-breakdown} to illustrate 
the power-consumption breakdown. The first 
three scenarios fix the P part by manually controlling 
the robot with specific revolutions per minute (RPMs), while
the fourth scenario covers the case when the robot is controlled 
by the navigation stack. For each scenario, we collect the power 
consumption by each of sensors, motors, and Jetson board. 
The breakdown of the board's power consumption 
(CPU, GPU, SOC, etc.) is read from \textsf{tegrastat}. 

\begin{table}[!htp]
\caption{Four scenarios used for power breakdown.}
\label{tab:power-dissipation-breakdown}
\resizebox{\columnwidth}{!}{
\begin{tabular}{|c|c|c|c|c|}
\hline
\textbf{Scenario} & \textbf{Motor RPM} & \textbf{Control} & \textbf{Navigation Stack} & \textbf{Detection} \\ \hline
1 & 1000 & Manual     & -                         & -                         \\ \hline
2 & 1000 & Manual     & \Checkmark & \Checkmark \\ \hline
3 & 2000 & Manual     & \Checkmark & \Checkmark \\ \hline
4 & [0, 1000] & Autonomous & \Checkmark & \Checkmark \\ \hline
\end{tabular}
}
\end{table}


\begin{figure}[!htp]
    \centering
    \includegraphics[trim=0cm 0cm 0cm 0cm, clip, width=\columnwidth]{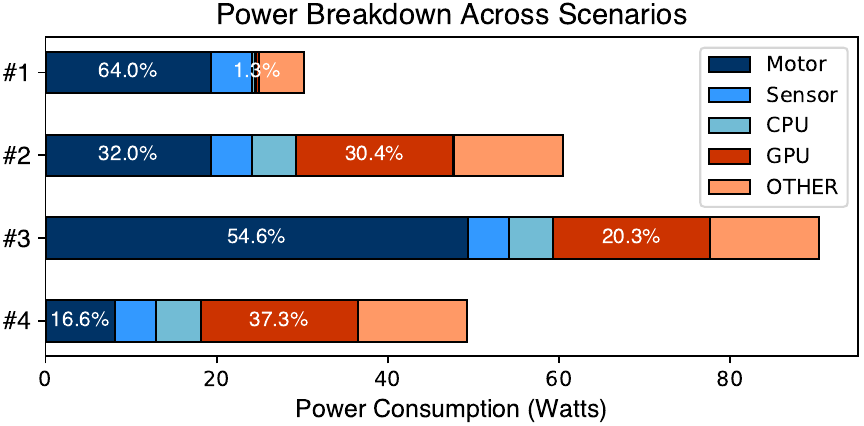}
    \caption{Power-consumption breakdown.}
    \label{fig:power-breakdown}
\end{figure}

Fig.~\ref{fig:power-breakdown} shows the breakdown 
of e2e power consumption in the four scenarios. 
The robot's total power consumption is found to range from 
30W to 90W, while the percentage of the motor's power consumption
ranges from 16.6\% to 54.6\%. The difference between scenario 1 
and 2 indicates that robot navigation and object detection 
consume 30W. Besides, the C subsystem dissipates more power 
than sensors and motors, whereas the GPU consumes more than
30.4\% of the total power. 
Comparing scenarios 2 and 3 shows that the motor's RPM increases 
from 1000 to 2000, increasing the motor's power consumption
by more than 22\%. For scenario 4, the navigation 
task controls the robot. 
The motor's RPM is set to be within [0, 1000]. The result for 
scenario 4 shows that the total power consumption is less than 
that of scenario 2 because the robot is not always moving. 
The navigation stack controls the robot to follow the planned
trajectories step-by-step to ensure navigation accuracy and safety. 
Motors are found to consume 16.6\% of the total power, 
while GPU consumes over 37.3\%. The variability in power 
consumption across C and P subsystems corroborates the 
importance of e2e optimization of power consumption.


\vspace{0.5mm}
\noindent\textbf{Observation 1:} \textit{
Power consumption varies with the scenario due to the different 
underlying control and navigation strategies. 
Real-time and fine-grained power prediction is, therefore,  essential 
for optimizing e2e power efficiency for AMRs.}





\subsection{Environment-Dependent Navigation Locality}

AMR's Navigation stack tends to process every sensory input 
from camera and LiDAR equally, which can be computationally expensive 
and inefficient. In reality, however,  not every pixel in each frame
is equally important for AMR's navigation.
While indoor AMRs are running at low speeds, we 
observe significant spatial and temporal localities 
for AMR's navigation. 
Below we show navigation localities with a timing 
analysis of sensor data view overlap and linear/angular 
control of the robot.

\begin{figure}[!htp]
    \centering
    \includegraphics[trim=0cm 0cm 0cm 0cm, clip, width=\columnwidth]{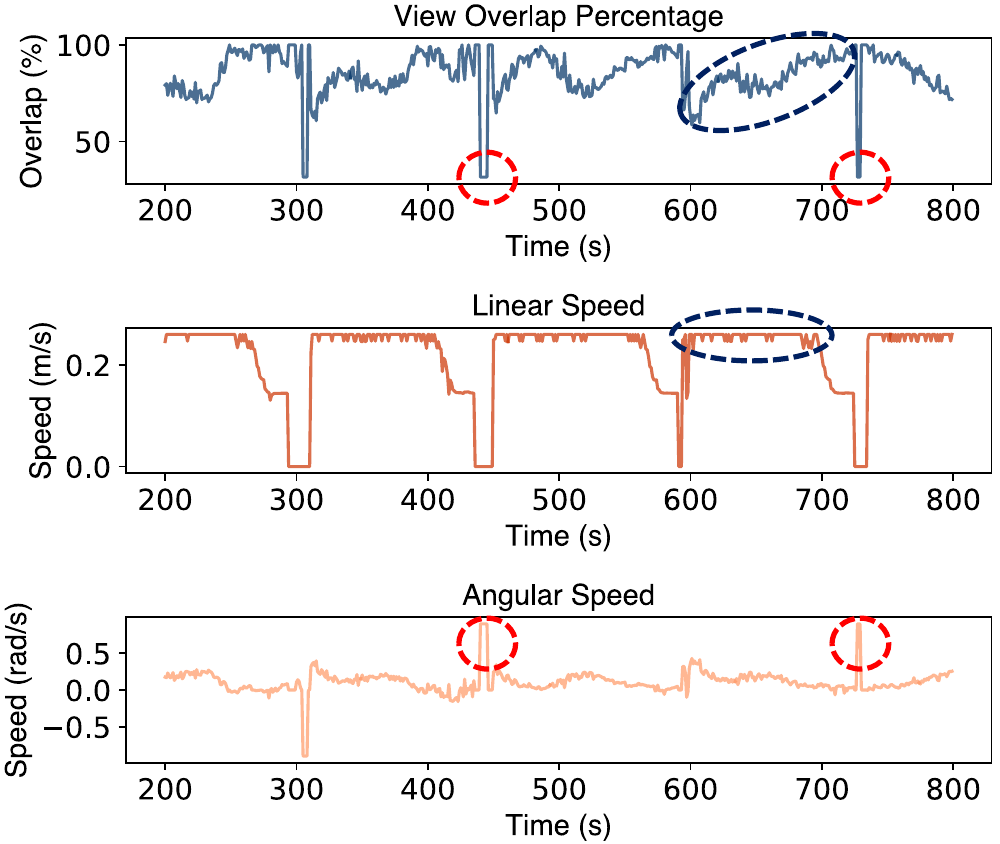}
    \caption{View overlap with linear/angular speed.}
    \label{fig:fov-sim-timeline}
\end{figure}

When an AMR is running, two consecutive point clouds or 
camera frames show a non-negligible similarity. 
To demonstrate its locality and correlation with the 
robot's movement, we recorded the sensor readings and 
calculated the view overlap percentage for two 
consecutive frames along with the robot's linear and 
rotational movements/speeds while running the Navigation stack. 
As illustrated in Fig.~\ref{fig:fov-sim-timeline}, for a 
significant portion of time, the overlap percentage
exceeds 80\%. Moreover, there exists a noticeable link 
between movement speed and overlap percentage. 
Sharp rotational movements result in substantial 
decreases in view overlap (marked as red circles), 
whereas the percentage varies minimally during steady 
linear motion (marked as blue circle). These findings 
underscore the substantial spatial and temporal 
relevance of sensing in the AMR's operation, especially 
when it avoids sharp rotational movements. This 
observation suggests the dynamic adjustment of AMR 
operations, such as ROS navigation and 
object detection, to save energy by leveraging
environment-dependent navigation localities.


\vspace{0.5mm}
\noindent\textbf{Observation 2:} \textit{
The AMR often operates at low speeds, enhancing both 
temporal and spatial localities. By exploiting these 
localities via dynamic adjustments of the navigation stack, 
a significant improvement of power-efficiency can 
be made.}

\subsection{Coordination of C and P Subsystems}
\begin{figure}[!htp]
\centering
\includegraphics[trim=0cm 0cm 0cm 0cm, clip, width=\columnwidth]{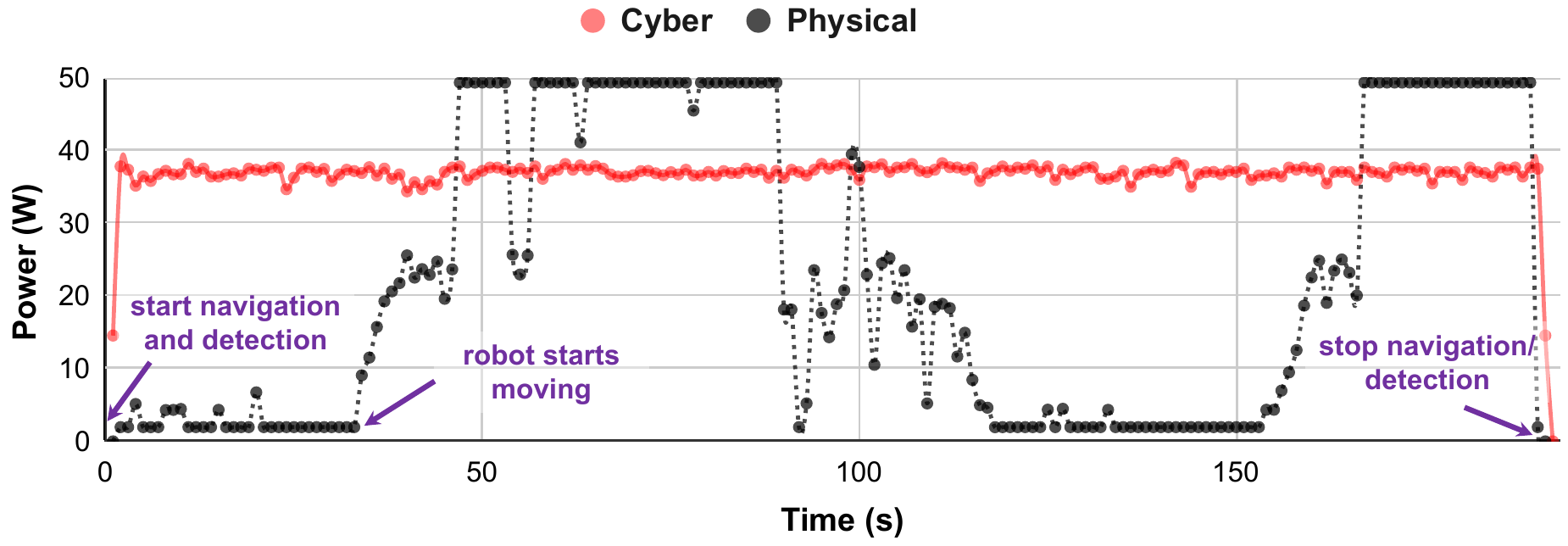}
\caption{Power timeline for AMR.}
\label{fig:AMR-power-timeline}
\end{figure}

According to the AMR's modular design, its C and 
P subsystems operate independently of each other for 
environment perception and robot control. This design lacks 
the coordination of C and P subsystems necessary to achieve 
optimal power-efficiency. To address this deficiency, we begin 
with a timing analysis to uncover the causes of power-inefficiency. 
Fig.~\ref{fig:AMR-power-timeline} shows the power 
consumption timeline for the AMR's computer, sensors, 
and motors.

For the C subsystem (computer and sensor), we observe 
that power consumption spikes to a high level when detection 
starts and remains high until the detection task ends. In contrast, 
the motors exhibit more variations in power consumption. When the 
navigation kicks off, power consumption remains at a low 
level and increases when the robot starts moving. The motor 
first accelerates to its highest speed, then maintains this 
speed, and finally decelerates to stop. When the robot 
stops, both navigation and detection cease, leading to a 
noticeable decrease in power consumption for both C 
and P subsystems.

\begin{figure}[!htp]
\centering
\includegraphics[trim=0cm 0cm 0cm 0cm, clip, width=\columnwidth]{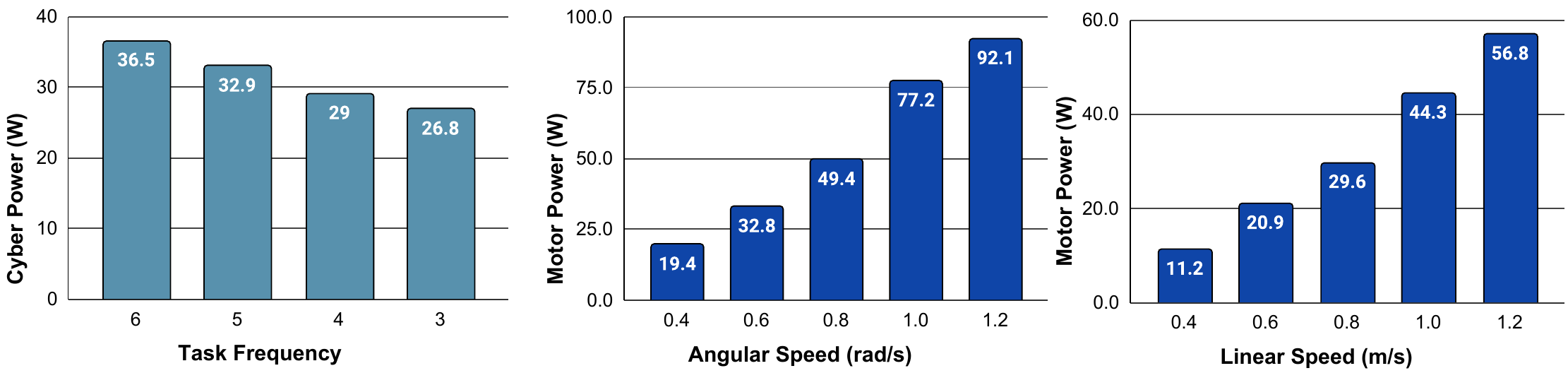}
\vspace{-4mm}
\caption{Impact of Task frequency and motor controls on 
C/P subsystem's power consumption.}
\label{fig:AMR-power-control}
\end{figure}

This timeline clearly shows the uncoordinated management 
of the two subsystems results in significant power wastage. 
First, even when the robot is not moving, the C subsystem continues 
to run at high power because it is responsible for environment 
perception while running navigation and detection tasks  
at a fixed frequency. If the C subsystem were aware of the robot's 
control state from the P subsystem, a significant amount of power 
could have been saved by dynamically changing the opearting 
frequency for running navigation and detection tasks. 
Fig.~\ref{fig:AMR-power-control} shows the effect of task frequency 
and motor controls on power cnosumption. 
The C-subsystem's power consumption can be reduced from 36.5W to
26.8W for YOLOv3 inference on Jetson AGX when the task 
frequency is lowered from 6Hz to 3Hz. Conversely, if the perception 
results were shared with and used by the P subsystem, 
a more power-efficient control trajectory could have been 
identified and executed (see in \S\ref{subsec:power-path-planner}), 
saving a considerable amount of power for the P subsystem. 
For example, reducing angular speed from 1.2 to 0.4 rad/s 
reduces power consumption from 92.1W to 19.4W.


\noindent\textbf{Observation 3:} \textit{
Uncoordinated C and P subsystems cause substantial power waste. 
To optimize power-efficiency, C and P subsystems should be 
jointly designed to achieve optimal power-efficiency.}

%% file: contents/4_System_Design.tex
\label{sec:system}
This section presents \name, which is a power-management 
system for AMRs. We first provide an overview of the 
proposed system, followed by a discussion of how \name addresses 
the aforementioned technical challenges. Then, we provide the 
details of \name's components.

\vspace{-2mm}
\subsection{System Overview}

\begin{figure}[!htp]
	\centering
	\includegraphics[width=\columnwidth]{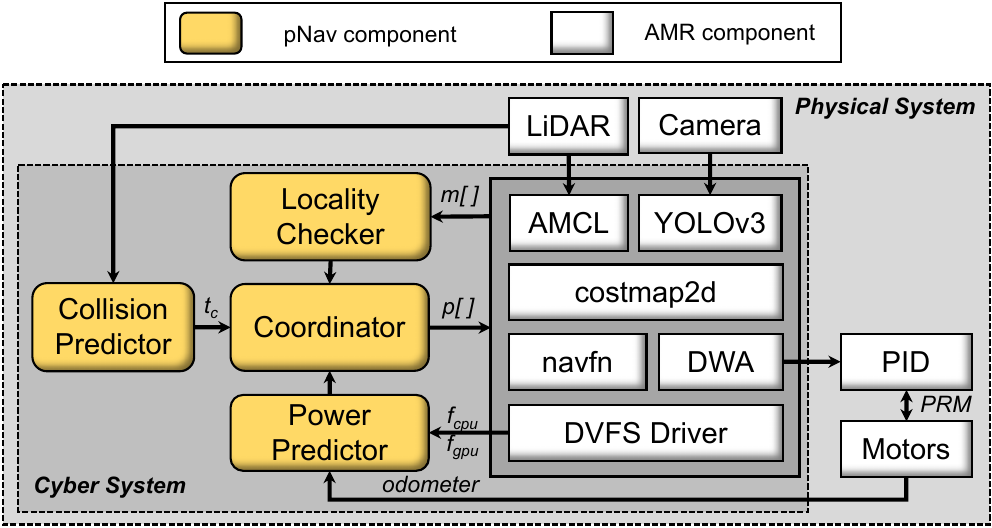}
	\caption{System architecture of \name.}
	\label{fig:robo-ems-design}
\end{figure}

Fig.~\ref{fig:robo-ems-design} depicts an overview of 
\name\ which is a comprehensive power-management system 
designed to enhance the energy efficiency and safety of 
AMRs. It consists of four key components: the Power 
Predictor, the Locality Checker, the Collision Predictor, 
and the Coordinator. The Power Predictor 
estimates the AMR's overall power consumption by 
analyzing motor RPM commands and feedback, as well as 
monitoring the current CPU/GPU frequencies. The 
Locality Checker oversees the navigation locality 
through monitoring AMCL particle's confidence, camera's 
FOV overlap, and path planning status. The Collision 
Predictor utilizes LiDAR data to forecast potential 
collisions and time-to-collision (TTC), 
crucial for safe navigation. 
Central to \name, the Coordinator synthesizes information 
from all modules to make informed decisions about CPU 
and GPU frequencies, and dynamically updates configurations 
for object detection, costmap, planning, and localization 
tasks based on TTC and power consumption. 
This holistic approach positions \name\ as an effective 
solution for managing power consumption while 
maintaining high safety for AMRs.

\vspace{-4mm}
\subsection{Technical Challenges}
\label{subsec:tech-challenges}

\name\ addresses several key technical challenges in 
creating an energy-efficient AMRs system.

\vspace{1mm}
\noindent\textbf{C1: Balancing End-to-End (e2e) Power 
Consumption:} This challenge deals with the variation of 
total power usage of the entire system, including 
sensors, computing devices, actuators, and other components. 
\name\ tackles this by introducing an e2e power predictor 
(Section~\ref{subsec:power-predictor}), which forecasts 
the power consumption of the whole AMR system and integrates 
it into the path planner.

\vspace{1mm}
\noindent\textbf{C2: Modeling Navigation Locality:} 
Utilizing navigation localities is key to achieving 
power-efficiency. However, achieving a lightweight and efficient 
model of navigation localities is non-trivial. \name\ addresses 
this challenge by introducing the Locality Checker which models 
and monitors the temproal and spatial localities across three 
leveles: sensor data, robot location, and robot trajectories.

\vspace{1mm}
\noindent\textbf{C3: Coordination of C and P Subsystems:} 
Ensuring the safety and efficiency of AMRs in dynamic 
and unpredictable environments presents significant challenges. 
\name\ tackles these challenges with a centralized Coordinator 
that continuously adjusts collision prediction and the real-time 
performance of localization, path planning, and robot control 
systems. This ensures both power-efficiency and 
timing/functional predictability.

\subsection{Power Predictor}
\label{subsec:power-predictor}

The total power consumption of the AMR 
($P_{\text{AMR}}$) is defined as the sum of the total 
power consumptions of the motors ($P_{M}$) and the  
cyber/embedded system ($P_{\text{ES}}$):
\begin{equation}
    P_{\text{AMR}} = P_{M} + P_{\text{ES}}.
\end{equation}

\subsubsection{Motor Power Consumption}

\begin{figure}[!htp]
	\centering
	\includegraphics[width=\columnwidth]{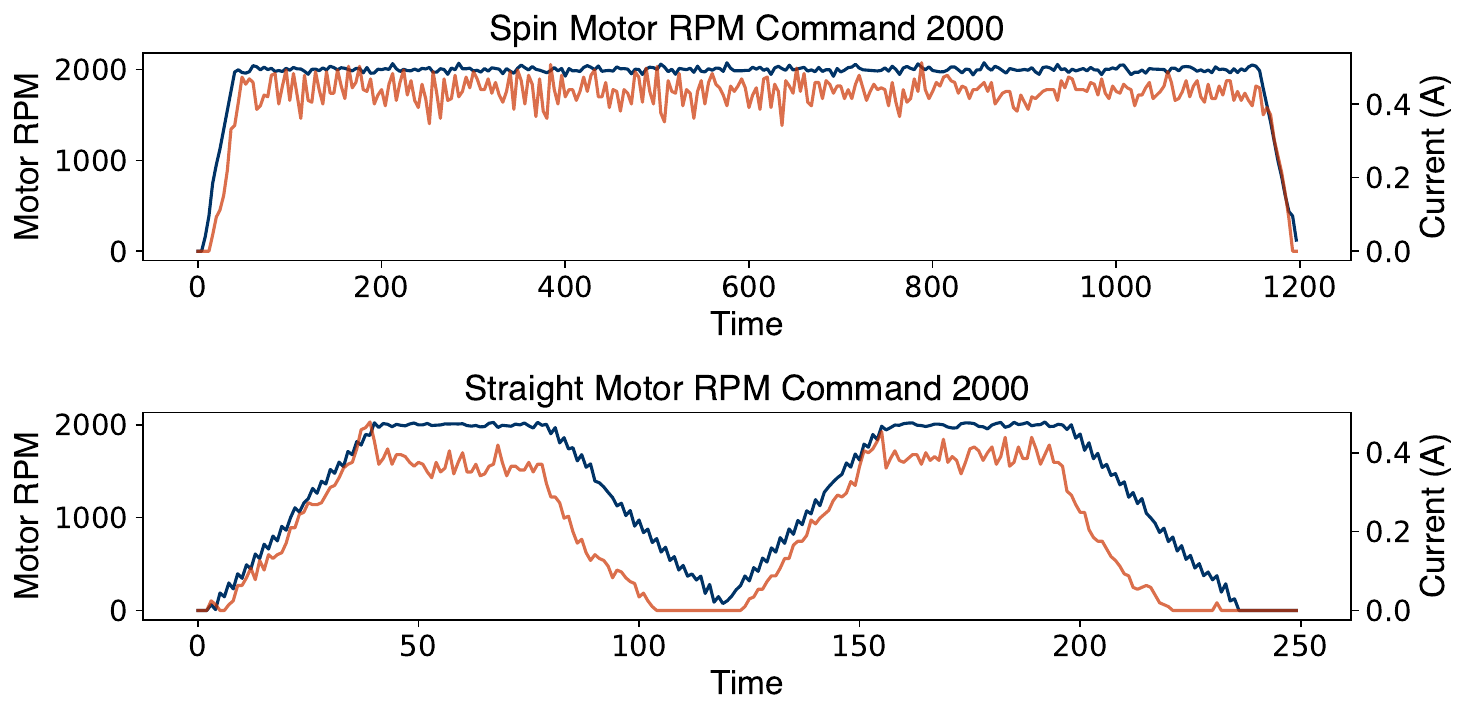}
	\caption{The timeline for motor's RPM and current.}
	\label{fig:motor-current-timeline}
\end{figure}

From the timline in Fig.~\ref{fig:motor-current-timeline}, 
we can observe a clear relationship between the motor's 
current and RPM.
The power consumption of the motors is given by:
\begin{equation}
    P_{M} = P(v_i, \omega_i, v_{i-1}, \omega_{i-1}),
\end{equation}
where $P(v_i, \omega_i, v_{i-1}, \omega_{i-1})$ is the 
power consumption of the motors at velocities $v_i$ and 
$\omega_i$ and previous velocities $v_{i-1}$ and 
$\omega_{i-1}$. To predict power consumption, we implement a 
multi-layer perceptron (MLP). The MLP model uses motor 
control commands and feedback as input parameters, aiming 
to predict current in real time at a millisecond level, 
a significant enhancement over the SOTA that relies on 
average current at a second level. 
The models' effectiveness is assessed using Mean Squared 
Error (MSE) as a loss function to be minimized.

\subsubsection{Power Consumption by Cyber/Embedded Subsystems}

The power consumption of a cyber/embedded system is:
\begin{equation}
    \label{equ:embedded-power}
    P_{\text{ES}} = F_1(f_{\text{cpu}}) + F_2(f_{\text{gpu}}) + F_3(f_{\text{cpu}}, f_{\text{gpu}}),
\end{equation}
where \( P_{\text{ES}}\) denotes the system's total 
power consumption. This equation consists of three primary 
components. \( F_1(f_{\text{cpu}}) \) calculates the power 
consumption by the CPU. The function \( F_1 \) depends 
on the CPU frequency \( f_{\text{cpu}} \). 
\( F_2(f_{\text{gpu}}) \) models the GPU's power usage 
where \( F_2 \) is a function of the GPU frequency 
\( f_{\text{gpu}} \). 
\( F_3(f_{\text{cpu}}, f_{\text{gpu}}) \) is responsible 
for calculating the power consumption of other peripherals 
in the system. 
The function \( F_3 \) is influenced by both CPU and GPU 
frequencies, capturing the observed behavior that 
peripheral power consumption jumps when the CPU frequency 
exceeds 2.0GHz and increases linearly with the GPU 
frequency.

\subsubsection{Integration into Path Planner}
\label{subsec:power-path-planner}

The path planner is designed for real-time collision avoidance 
and trajectory generation for mobile robots~\cite{ros-dwa-planner}.
It operates by generating a set of velocity commands 
considering the robot's dynamics, kinematics, and current 
environmental context. In general, the path planner evaluates 
these potential commands based on such criteria as collision 
avoidance, heading toward the goal, and 
velocity~\cite{fox1997dynamic}. 
\begin{align}
\label{equ:dwa-with-power}
G(\mathbf{v}, \omega) &= \sigma(\alpha \cdot \text{heading}(\mathbf{v},\omega) + \beta \cdot \text{dist}(\mathbf{v}, \omega) \nonumber \\
&\quad + \gamma \cdot \text{vel}(\mathbf{v}, \omega)
+ \theta \cdot p(\mathbf{v}, \omega)).
\end{align}
\name\ integrates the power prediction into the cost function 
of path planner as shown in Eq.~(\ref{equ:dwa-with-power}). 
The added cost function $\theta \cdot p(\mathbf{v}, \omega)$ 
is the predicted motor power consumption 
given the velocity and rotation for each sample. 
For all sampled velocity commands, \name\ selects the optimal 
command that maximizes these criteria while ensuring safety.

\subsection{Locality Checker}
\label{subsec:locality-checker}

To handle a dynamic environment and make the AMR system 
energy-efficient, we propose a novel approach to 
model the AMR navigation locaility at three levels:
data level (robot FOV overlap monitor), position level 
(particle confidence ratio), and path level 
(trajectory waypoint monitor).

\subsubsection{Robot FOV Overlap Monitor}

\begin{figure}[!htp]
    \centering
    \includegraphics[trim=0cm 0cm 0cm 0cm, clip, width=.85\columnwidth]{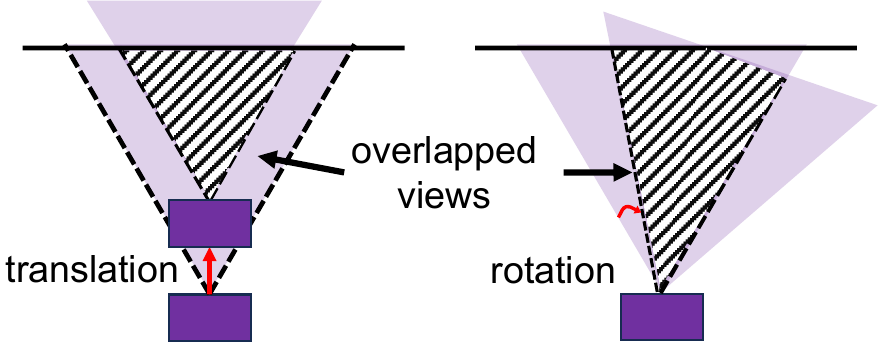}
    \caption{View overlap when robot is moving.}
    \label{fig:fov-overlap}
\end{figure}

Data similarity in sensing reflects the navigation locality 
at the data level. The Field of View (FOV) of a camera defines 
the visible extent of the environment through the camera lens 
at a particular instance. This angular measurement of the 
captured scene is crucial in determining the breadth of 
the environment that can be visualized and recorded. 
In the context of AMRs, a significant FOV similarity is 
often observed in consecutive frames, particularly when 
the AMR is stationary during the planning phase. As a 
result, the AMR's FOV remains largely unchanged when it 
is not moving. However, the FOV can vary when the AMR is 
in motion. Consequently, if a high degree of overlap in 
the FOV between successive frames is detected, we prefer
the reduction of the frequency of object detection 
tasks, thus improving power-efficiency.


\begin{figure}[htbp]
  \resizebox{\columnwidth}{!}{%
    \begin{minipage}{\columnwidth}
      \begin{algorithm}[H]
        \caption{Calculating Camera FOV Overlap Ratio.}
        \label{alg:fov_overlap}
        \begin{algorithmic}[1]
          \small
          \State \textbf{Input:} robot positions $p[t_1]$, $p[t_2]$, robot orientations $o[t_1]$, $o[t_2]$, resolution $w$, $h$, focal length $f_x$, $f_y$
          \State \textbf{Output:} FOV overlap ratio $\theta$

          \Function{FOVOverlap}{$p[t1]$, $o[t1]$, $p[t2]$, $o[t2]$, $w$, $h$, $f_x$, $f_y$}
              \State $FOV_w \gets 2\arctan\left(\frac{w}{2 \cdot f_x}\right)$, $FOV_h \gets 2\arctan\left(\frac{h}{2 \cdot f_y}\right)$ 
              \Comment{Horizontal and Vertical FOV}

              \State $Fru_{t_1} \gets \textbf{FOV\_polygon}(p[t_1], o[t_1], FOV_w, FOV_h)$ \Comment{Frustum intersection at $t1$}
              \State $Fru_{t_2} \gets \textbf{FOV\_polygon}(p[t_2], o[t_2], FOV_w, FOV_h)$ \Comment{Frustum intersection at $t2$}

              \State $area_{t_1} \gets \textbf{Area}(Fru_{t_1})$
              \State $area_{t_2} \gets \textbf{Area}(Fru_{t_2})$

              \State $overlapArea \gets \textbf{Area}(Fru_{t_1} \cap Fru_{t_2})$ \Comment{Intersection area}

              \State $\theta_{t_2} \gets \frac{overlapArea}{\min(area_{t_1}, area_{t_2})}$

              \State \textbf{return} $\theta_{t_2}$
          \EndFunction
        \end{algorithmic}
      \end{algorithm}
    \end{minipage}%
  }%
\end{figure}

To quantify the similarity between two camera views during robot 
motion, we introduce an FOV overlap ratio that leverages the 
camera's intrinsic parameters and the robot's movement. As shown in 
Fig.~\ref{alg:fov_overlap}, the overlap area can vary depending 
on the robot's translation and rotation. Algorithm~\ref{alg:fov_overlap}
outlines the computation steps for the FOV overlap ratio. 
The process starts by initializing the robot’s positions, orientations, 
and camera parameters (Line 3). The horizontal and vertical FOV 
angles are computed from the camera intrinsics (Line 4), followed by 
generation of the FOV polygons (frustums) for the initial and final poses 
by projecting the camera's view frustum into the environment based on its 
pose and FOV angles (Lines 5–6). The areas of the individual frustums are 
then calculated (Lines 7–8), typically using a polygon area function 
to account for the shape of the projected view. 
Their intersection area—--representing the region visible in both camera 
views—--is determined by computing the geometric overlap between the two frustums 
(Line 9). Finally, the FOV overlap ratio $\theta$ is computed by normalizing 
the intersection area with respect to the smaller of the two individual frustum 
areas (Line 10), providing a scale-invariant measure of visual consistency 
across the robot’s motion.


\subsubsection{Particle Confidence Ratio}

Localization confidence reflects the navigation locality at 
the position level. This metric is pivotal in gauging 
the precision and reliability of the robot's localization 
algorithm~\cite{ros-amcl}. A collection of potential 
positions, termed {\em particles}, are established and dynamically 
refined based on incoming scan and odometry data. The particle 
that accumulates the highest scores undergoes clustering to 
determine the AMR's final position. In this approach, 
the variation of particles serves as an indicator of 
the localization algorithm's confidence level. The particles' 
variations are small when the SLAM system is confident about its 
localization. We leverage this insight and propose a new metric, 
called {\em confidence ratio}, that involves analysis of the 
variances of the potential locations (particles). The positions 
of the particles are represented in a matrix form, where \( x_i \) 
and \( y_i \) correspond to the coordinates of each particle:
\vspace{-1mm}
\begin{align}
\text{positions} = \left\{ (x_1, y_1), (x_2, y_2), \ldots, (x_n, y_n) \right\}
\end{align}

The variance of these positions is then calculated in 
both the $x$ and $y$ dimensions, capturing dispersion:
\begin{align}
    \text{variance} &= 
    \begin{bmatrix}
        \text{Var}(x) \\
        \text{Var}(y) \\
    \end{bmatrix}
    =
    \begin{bmatrix}
        \frac{1}{n}\sum_{i=1}^{n}(x_i - \bar{x})^2 \\
        \frac{1}{n}\sum_{i=1}^{n}(y_i - \bar{y})^2 \\
    \end{bmatrix}
\end{align}
Finally, the confidence ratio is computed as the inverse 
of the sum of these variances, providing a quantifiable 
metric of localization confidence:
\begin{align}
    \text{confidence\_ratio} &= 
        \frac{1}{\text{Var}(x) + \text{Var}(y)} 
\end{align}

Since particle/hypothesis-based approaches are common in SLAM 
systems, they offer an effective and correct way to assess 
the accuracy of SLAM localization. This method is used to monitor 
the status of SLAM localization and navigation locality.


\subsubsection{Trajectory Waypoint Monitor}

Trajectory waypoints from the path planner reflect the navigation 
locality at the path level. In general, the path planner periodically 
generates global and local paths. While the environment can be dynamic 
and unpredictable, we observe that the waypoints for both local and 
global paths can be predicted within a certain time window (200ms). 
This is because the planner usually sets maximum linear and angular 
speed which constrains AMR's movement within a certain 
range~\cite{ros-navigation}.

Based on this insight, we designed the trajectory waypoint monitor 
to track the length of both the global and the local plans. The 
monitor's key functionality lies in its ability to compare the 
current plan lengths with their previous states. If the global 
plan's length decreases without the corresponding increase in the 
local plan's length, it suggests that the robot is successfully 
moving toward its goal. In contrast, an increase in the 
global plan's length could indicate a deviation from the 
intended path or a lack of progress. 


Running at a regular interval, the trajectory waypoint monitor 
continuously invokes its analysis, providing real-time 
feedback on the robot's path planning efficacy. This 
continuous monitoring is crucial for quickly identifying  potential 
navigation inefficiencies, thus enhancing the overall performance 
of AMRs in dynamic environments. 

\vspace{-3mm}
\subsection{Collision Predictor}
\label{subsec:collision-predictor}

The collision predictor is adept at real-time monitoring 
for potential collisions in dynamic environments and  
composed of two primary components: the timeline analyzer 
and the calculation of time to collision (TTC).

\subsubsection{Timeline Analyzer}

The timeline analyzer is responsible for predicting 
the e2e delays in AMR's tasks. 
Based on the software pipeline depicted in 
Fig.~\ref{fig:donkey-software-pipeline}, the localization 
results from SLAM and the detection results from the detector 
are sent to the local cost-map. This cost-map is then updated 
and forwarded to the local planner. Finally, the path 
planner generates local plans and transmits them to the AMR.

\begin{figure}[!htp]
	\centering
	\includegraphics[width=.8\columnwidth]{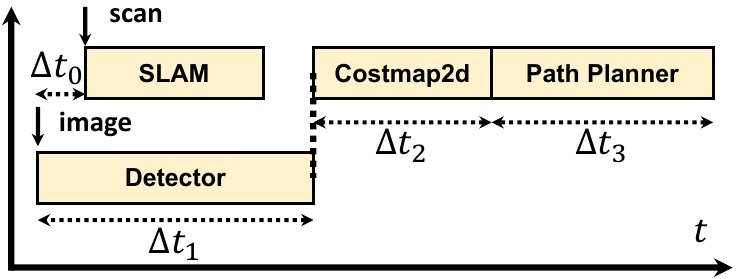}
	\caption{Timeline analysis of the AMR tasks.}
	\label{fig:AMR-timeline}
\end{figure}


Fig.~\ref{fig:AMR-timeline} illustrates the timeline 
of AMR's tasks. The global planner is excluded from this 
analysis, as it is typically executed once per navigation. 
Given that the camera and LiDAR operate at 
different frequencies, there is an initial delay, 
$\Delta t_0$. The delay $\Delta t_1$ is the maximum of 
SLAM's or detector's delay, with SLAM's delay 
$t_{slam}$  (detector's delay $t_{object}$), depends on 
CPU (GPU) frequency. 
$\Delta t_2$ represents the delay for costmap2d updates, 
while $\Delta t_3$ is the delay for the local planner, 
with both primarily influenced by CPU frequency.

\subsubsection{Time-to-Collison}

The Time-to-Collision (TTC) component of the collision 
analyzer is designed for the safety of AMRs, utilizing 
real-time LiDAR data and the robot's current motion. 
It calculates the time until a potential collision by 
dividing the shortest distance (\( f_d \)) from the LiDAR scan 
gives a drivable window ($V_{max}, W_{max}$). 
This approach offers a simple yet effective estimation of 
the urgency to avoid collisions, ensuring the robot 
dynamically adapts its path in changing environments. 
If no obstacle is detected (\( f_d = \infty \)), the TTC 
is set to $\infty$, indicating a clear path ahead.
\begin{align}
    f_d &= \textbf{min} (\text{Scan}(d + v_i \cdot t, \theta + \omega_i \cdot t)) \\
\text{subject to: } & 0 \leq v_i \leq V_{\text{max}}, -W_{\text{max}} \leq \omega_i \leq W_{\text{max}} \notag \\
    t_c &= 
    \begin{cases}
        \frac{f_d}{V_{max}}, & \text{if } f_d \neq \infty \\
        \infty, & \text{otherwise}
    \end{cases}
\end{align}

To this end, we propose a safe time metric which 
quantifies the available time before a potential 
collision by comparing the TTC with the aggregate 
delays in the robot's operational timeline. 
This metric is defined as:
\begin{align}
    \label{equ:safe-time}
    s_t = t_c - (\Delta t_0 + \Delta t_1 + \Delta 
    t_2 + \Delta t_3),
\end{align}
where \( t_c \) represents the TTC, determined from 
the real-time LiDAR data and the robot's velocity. 
The components \( \Delta t_0, \Delta t_1, \Delta t_2, 
\Delta t_3 \) signify delays associated with the camera and 
LiDAR data processing, SLAM and detector computations, 
costmap updates, and local path planning by the path 
planner, respectively. This metric is scrutinized under 
two distinct scenarios: the average case, which employs 
average values of delays, and the extreme case, which 
accounts for the maximum observed delays. Notably, in 
extreme cases, these delays can be significantly 
influenced by the CPU/GPU frequencies, impacting the 
safe time metric. 

\subsection{Coordinator}
\label{subsec:coordinator}

The coordinator is designed to collect all profiling 
information from the power predictor, the navigation 
checker, and the collision predictor to determine the 
best system configurations that achieve the lowest 
total power consumption while guaranteeing the AMR's 
safety. The system configurations consist of the 
CPU/GPU frequencies and parameters selection for 
the navigation stack.

\subsubsection{CPU/GPU Frequency Optimization for AMRs} 

A key aspect of enhancing energy efficiency in AMRs
involves the strategic selection of CPU and GPU 
frequencies. This process aims to reduce the overall 
power consumption (\(P_{ES}\)) of the embedded system, 
while simultaneously ensuring that the TTC (\(t_c\)) 
is adequately greater than the worst-case sum of all 
operational delays (\(\Delta t_0 + \Delta t_1 + 
\Delta t_2 + \Delta t_3\)). Given that the operational 
frequency of the camera is 30Hz and the LiDAR is 5Hz, 
we limit the delay \(t_0\) to a maximum of 33ms. 
The delay \(t_1\) is dependent on both CPU and GPU 
frequencies (\(f_{cpu}\) and \(f_{gpu}\)), while \(t_2\) 
and \(t_3\) are influenced primarily by \(f_{cpu}\). 
Our analysis covers seven distinct CPUs and six GPU 
frequency levels. By offline profiling, we can identify the 
worst-case scenarios for each frequency pairing. 
The optimization problem is thus defined as that of 
minimizing \(P_{ES}\), ensuring that the effective 
TTC remains above a critical threshold (\(t_d\)):
\begin{equation}
\label{equ:optimization-model}
\begin{aligned}
& \textbf{min } P_{\text{ES}}(f_{\text{cpu}}, f_{\text{gpu}}) \textbf{ s.t. } t_c - (\Delta t_0 + \\
& \Delta t_1 (f_{\text{cpu}}, f_{\text{gpu}}) + \Delta t_2(f_{\text{cpu}}) + \Delta t_3(f_{\text{cpu}})) \geq t_d
\end{aligned}
\end{equation}

Considering a small number of possible combinations of CPU 
and GPU frequencies, the search for the 
optimal settings are both efficient and 
computationally economical.

\subsubsection{Navigation Coordination}

With the CPU and GPU frequencies determined and a safe 
response time (\(s_c\)) established for worst-case scenarios, 
the coordinator can dynamically toggle the AMR between 
"power-saving mode" and "recovery mode." This optimization 
balances performance and safety effectively.

\vspace{1mm}
\noindent\textbf{Power Saving Mode.} Initially, the coordinator 
assesses results from the Locality Checker to perform a 
locality test. Specifically, if the FOV overlap ratio and 
particle confidence fall within predefined thresholds and 
the path planner demonstrates progress, the AMR enters 
'power saving mode.' This mode adjusts the CPU and GPU 
frequencies to manage the embedded system dynamically. 
Additionally, the coordinator integrates a power predictor 
with the path planner (Eq.\ref{equ:dwa-with-power}) and 
applies a delay (\(s_t\)) to the detection task, guided 
by timeline analysis (Eq.\ref{equ:safe-time}).

\vspace{1mm}
\noindent\textbf{Performance Mode.} If the FOV overlap ratio 
or particle confidence drops below the set threshold, or 
if the path planner deviates from its intended path, the 
AMR switches to 'performance mode.' In this mode, the 
coordinator adjusts the navigation stack to facilitate 
recovery. To enhance the FOV overlap ratio, it resets 
the detection task's delay to zero. To boost particle 
confidence, it increases the number of particles and 
the scan beams for the SLAM algorithm. For deviations in 
path planning, it raises control frequencies and reduces 
sampling time to refine the trajectory.

%% file: contents/Donkey.tex
\label{donkey}

To address the challenge of limited comprehensive power profiling in AMRs, 
we have developed the Donkey power-profiling platform, an AMR equipped 
with current sensors on each subsystem (motors, sensors, 
computing) and specialized software for power profiling. 
Below we outline Donkey's hardware,
software pipeline, and power monitoring system.

\begin{figure}[t]
    \centering
    \includegraphics[width=.75\columnwidth]{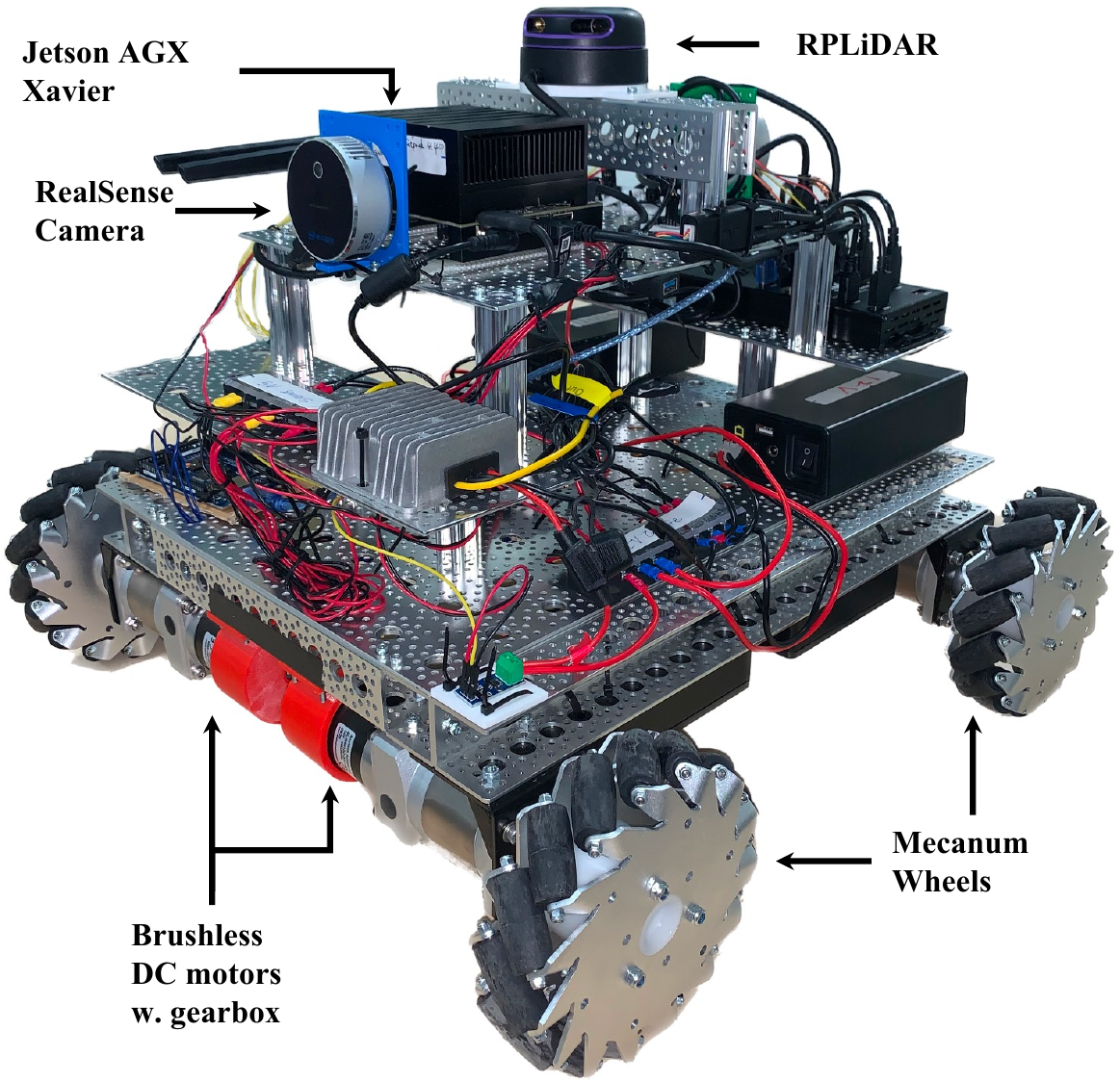}
    \caption{Donkey platform.}
    \label{fig:donkey-platform}
\end{figure}

\subsection{Hardware Design}


For representative power profiling, Donkey is designed and built to follow 
the general rules~\cite{mei2004energy, gaskell2024mbot, turtlebot-github}. 
Fig.~\ref{fig:donkey-platform} 
shows an overview of Donkey which are equipped with four Mecanum wheels. 
Four brushless DC motors with hall sensor and gearbox are used as 
the locomotion part. Each motor is controlled by a separate controller 
using a Pulse-width modulation (PWM) signal on serial communication links. 
Donkey is also equipped with  an Intel RealSense L515 camera and 
an RPLiDAR A3. Its compute board is an NVIDIA Jetson AGX Xavier. 
The sensors and motor controllers are connected to the Jetson
board through a 10-port USB hub.



\vspace{-3mm}
\subsection{Software Pipeline}

\begin{figure*}[t]
    \centering
    \begin{subfigure}[b]{0.63\textwidth}
        \centering
        \includegraphics[width=\columnwidth]{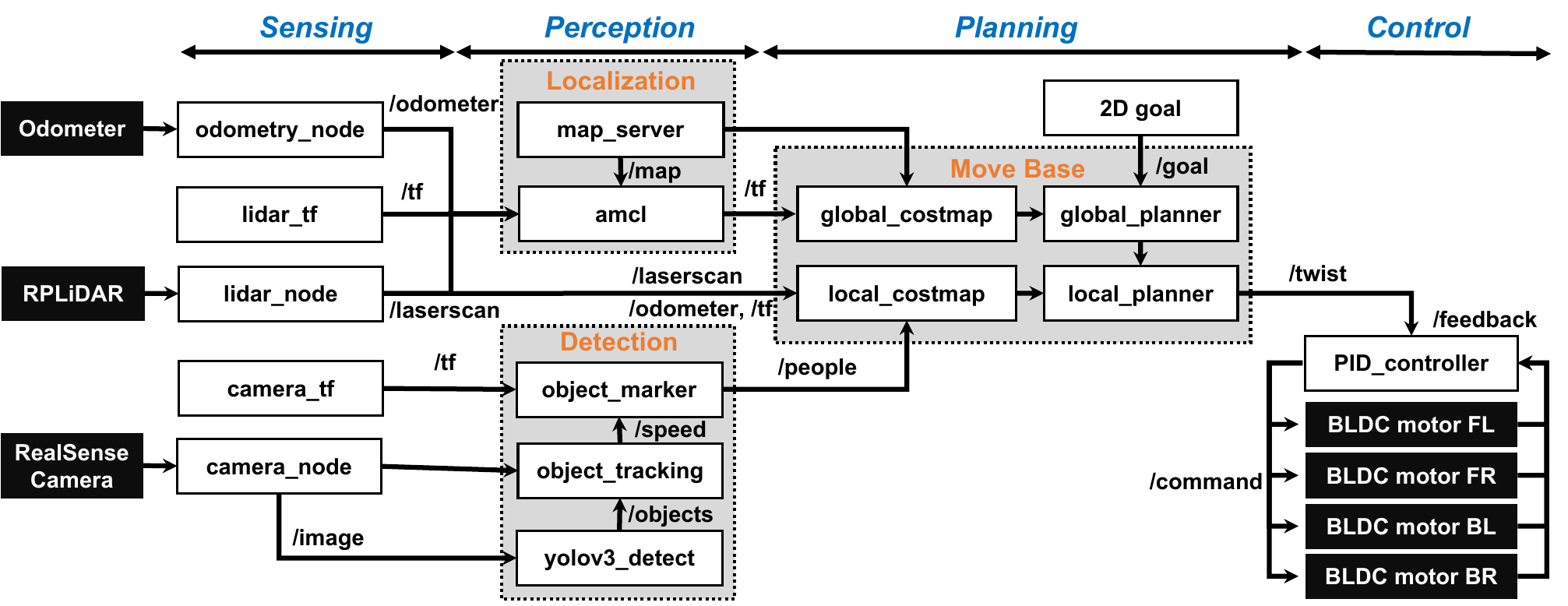}
        \caption{}
        \label{fig:donkey-software-pipeline}
    \end{subfigure}
    \hspace{1em} 
    \begin{subfigure}[b]{0.33\textwidth}
        \centering
        \includegraphics[width=\columnwidth]{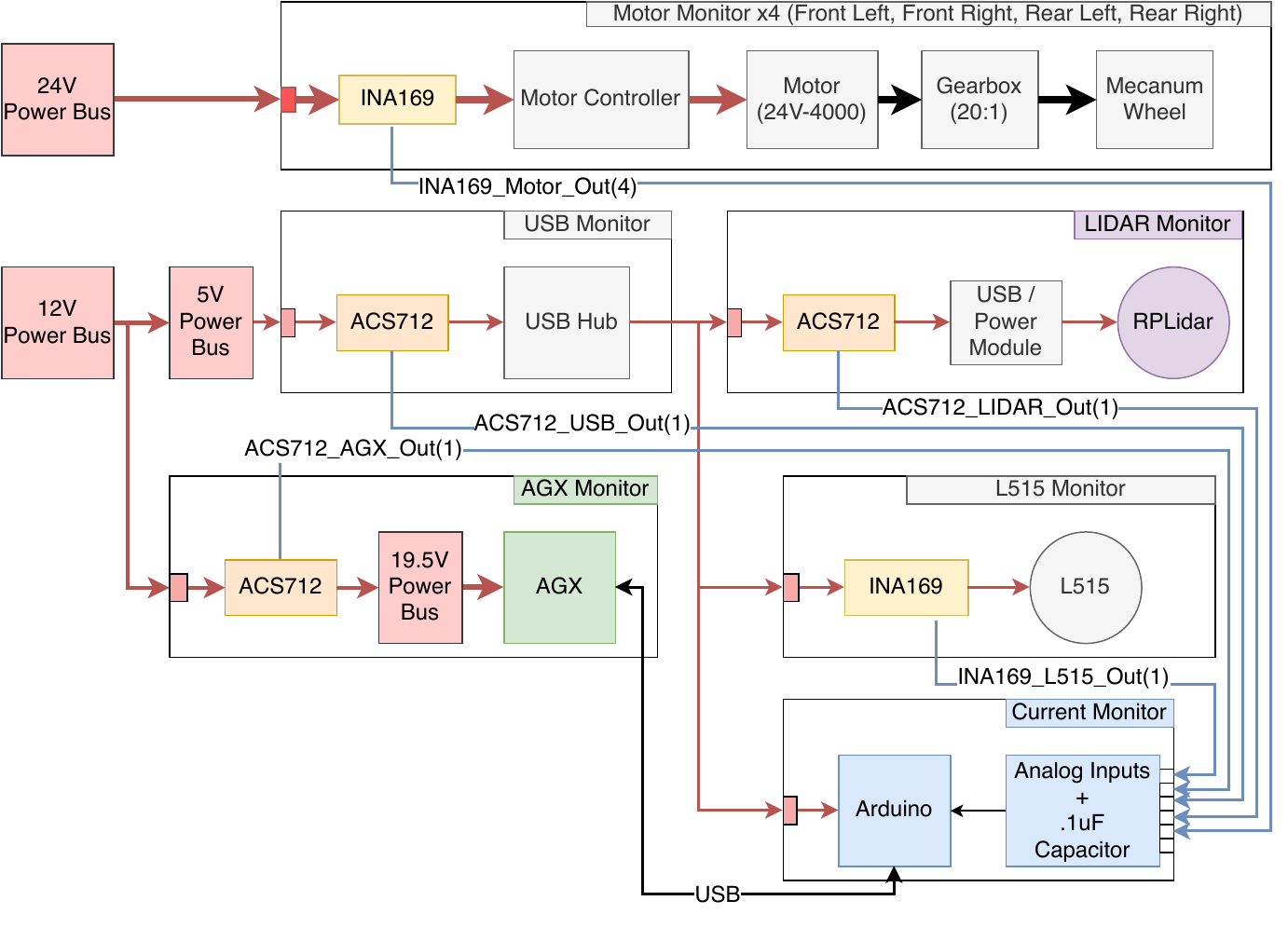}
        \caption{}
        \label{fig:donkey-pds}
    \end{subfigure}
    \caption{Overview of (a) Donkey software pipeline and (b) power 
    distribution and monitoring system.}
    \label{fig:donkey-combined}
\end{figure*}

The autonomous driving software pipeline on Donkey is built with ROS, 
a middleware framework that supports modular robot development 
through nodes and topics. The pipeline consists of 4 main stages
as shown in Fig.~\ref{fig:donkey-software-pipeline}: 
sensing, perception, planning, and control. 
Each component is implemented using ROS nodes (individual processes) 
and topics (communication channels).

\vspace{1mm}
\noindent\textbf{Sensing:} Donkey uses an RPLiDAR and a RealSense camera 
for environmental sensing. Sensor data acquisition and publication are 
handled by dedicated ROS nodes: \textsf{lidar\_node} for the RPLiDAR 
and \textsf{camera\_node} for the camera. These nodes publish data in 
standard ROS message formats such as \textsf{LaserScan} and \textsf{Image}. 
An \textsf{odometer} node estimates the robot's movement using 
an omniwheel kinematic model. 

\vspace{1mm}
\noindent\textbf{Perception:} This stage enables the robot to localize 
itself and perceive its environment. A map of the indoor environment 
is first generated using the \textsf{gmapping} package, and localization is 
performed using Adaptive Monte Carlo Localization (AMCL) 
which uses odometer and LiDAR's scan to estimate the robot's pose. 
For dynamic environment perception, Donkey integrates LiDAR and camera data. 
The \textsf{yolov3\_detect} node performs object detection on camera images 
using a pre-trained YOLOv3 model. Detected bounding boxes are passed to the 
\textsf{object\_tracking} node, which combines them with depth images to 
track the object's movement. The \textsf{object\_marker} node then 
transforms the object's position and velocity into the robot's frame 
and publishes a \textit{/people} topic for the planning module.

\vspace{1mm}
\noindent\textbf{Planning:} This stage determines a path from the robot’s 
current location to a target position. Donkey’s planning system extends 
the ROS Navigation Stack to support sensor fusion between LiDAR and 
camera inputs. It uses a global planner based on Dijkstra’s algorithm 
to compute an overall route on a static costmap, derived from the map server. 
A local planner, \textsf{base\_local\_planner}, generates motion trajectories 
using the local costmap, which is continuously updated with dynamic obstacle 
data including the object's speed and direction.

\vspace{1mm}
\noindent\textbf{Control:} The control stage translates velocity commands 
into motor actions. Donkey uses 4 PID controllers to regulate the speed of 
its brushless DC motors connected to mecanum wheels. Each controller 
adjusts the motor's RPM based on the difference between the desired 
and actual speeds, calculated from hall sensor feedback. 
The control signal is generated by using a weighted sum of  proportional, 
integral, and derivative components, and sent as a \textit{/twist} 
message to the motors. This feedback loop ensures stable and desired 
robot movement in response to planning commands.

\subsection{Power Distribution and Monitoring System (PDMS)}

\noindent\textbf{Current Sensor-Based Power Monitoring.} 
The Power Distribution and Monitoring System (PDMS) on Donkey enables 
real-time current monitoring and control across all motors, 
compute units, and sensors using ACS712 (\(\pm1.5\%\) error) and 
INA169 (\(\pm2\%\) error) sensors, each supporting up to 5A of current 
\cite{allegro_microsystems_acs712_2021,texas_instruments_ina1x9_2021}. 
An Arduino Mega 2560 functions as an external ADC, filtering analog 
signals with 0.1\(\mu\)F capacitors and transmitting digitized data to 
the AGX platform via USB. The Arduino also manages relays to 
selectively enable or disable power to individual components. 

The PDMS supports three core functions: accurate current measurement 
(\(\pm2\%\) max error), relay-based power control, and a serial interface 
for data transmission and control commands. As shown in 
Fig.~\ref{fig:donkey-pds}, the 24V supply powers the motor controller 
through an INA169 sensor, while the 12V line feeds two monitored converters: 
one supplying 5V to a USB hub connected to sensors (RPLiDAR, 
Intel RealSense L515s, and the Arduino), and another delivering 19.5V to 
the AGX. All current data are published as real-time ROS messages 
for system-level monitoring and analysis.

\vspace{1mm}
\noindent\textbf{Software-based Power Monitoring.} In addition to 
hardware-based monitoring, the PDMS includes a software-based power 
monitor using \textsf{tegrastats} to track real-time power usage 
on the Jetson AGX board. 
The board features two 3-channel INA3221 monitors for key power rails. 
One monitor tracks \textsf{VDD\_GPU}, \textsf{VDD\_CPU}, 
and \textsf{VDD\_SOC}, while the other monitors \textsf{VDD\_CV}, 
\textsf{VDDRQ}, and \textsf{SYS5V}, with measurement errors under 5\%. 
\textsf{Tegrastats} reports component-level metrics, including 
current and average power in milliwatts, GPU and memory usage 
(e.g., \textsf{GR3D\_FREQ} for GPU utilization and \textsf{EMC\_FREQ} 
for memory bandwidth). These readings are integrated into ROS via the 
\textsf{ros\_jetson\_stats} package, publishing real-time system 
power data as ROS messages.

\begin{center}
\small
\fcolorbox{black}{gray!10}{\parbox{.9\linewidth}{
\textsf{[RAM 16496/31760MB SWAP 0/15880MB CPU [19\%@2264, 10\%@2264, 12\%@2264, 16\%@2264, 19\%@2265, 25\%@2265, 25\%@2264, 37\%@2265] EMC\_FREQ 9\%@2133 GR3D\_FREQ 45\%@318 GPU 930/930 CPU 2943/2943 SOC 2942/2942 CV 0/0 VDDRQ 774/774 SYS5V 3772/3772]}}
}
\end{center}

%% file: contents/5_Implementation.tex
\label{sec:implementation}

To evaluate the performance of \name, we integrate 
its implementation into the ROS Navigation stack~\cite{ros-navigation}.  

\begin{figure}[!htp]
	\centering
	\includegraphics[width=\columnwidth]{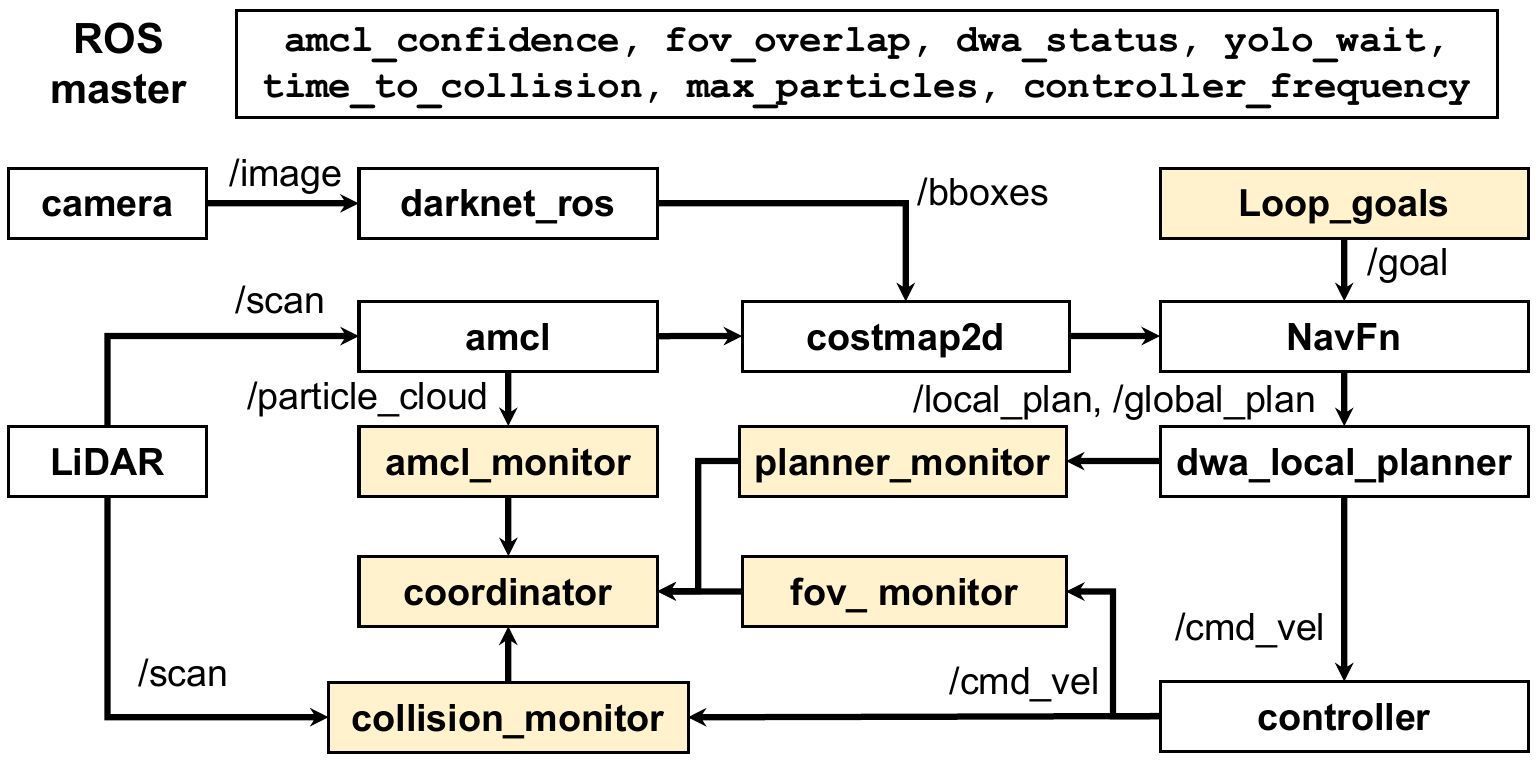}
	\caption{ROS Implementation of \name.}
	\label{fig:ros-implementation}
\end{figure}


\noindent\textbf{ROS Framework.} 
Fig.~\ref{fig:ros-implementation} show a 
comprehensive ROS framework for \name. The system 
integrates multiple components, each dedicated 
to specific aspects of AMR operations, and is 
coordinated through ROS. \texttt{darknet\_ros} node 
runs YOLOv3 for object detection, while 
\texttt{dwa\_local\_planner} and \texttt{NavFn} nodes 
are the global and local planners. The key elements of 
\name\ include specialized nodes for AMCL 
confidence estimation, planner path monitoring, FOV 
overlap calculation, and a central coordinating node 
that dynamically adjusts various configurations based 
on real-time AMR status monitoring.

\begin{table}[!ht]
\centering
\caption{ROS Parameters in \name.}
\label{tab:robo-ems-parameters}
\resizebox{\columnwidth}{!}{%
\begin{tabular}{|l|l|l|}
\hline
\textbf{Name}             & \textbf{Type}  & \textbf{Range}           \\ \hline
\texttt{amcl\_confidence}          & float          & [0.0, $\infty$]               \\ \hline
\texttt{fov\_overlap}              & float          & [0.0, 1.0)               \\ \hline
\texttt{dwa\_status}               & boolean        & \{PLANNING, RUNNING\}          \\ \hline
\texttt{yolo\_wait}                & integer (ms)   & [0, 200]            \\ \hline
\texttt{time\_to\_collision}       & float (seconds)& [0.0, $\infty$)          \\ \hline
\texttt{max\_particles}            & integer        & [500, 3000]            \\ \hline
\texttt{controller\_frequency}     & integer (Hz)   & [10, 20, 30]            \\ \hline
\end{tabular}
}
\end{table}

The \texttt{amcl\_monitor} node continuously assesses 
the localization accuracy by analyzing the variance 
of particle clouds, providing a metric of 
localization reliability. 
The \texttt{planner\_monitor} observes the lengths of 
global and local plans generated by the DWA planner, 
offering insights into the robot's progress towards 
its goal and its navigational efficiency. The 
\texttt{fov\_monitor} node plays a key role in 
optimizing the operation of vision-based systems like 
YOLO by determining the overlap between two consecutive 
frames, allowing for the adjustment of 
processing frequency to conserve power. Moreover, 
the \texttt{colision\_monitor} node takes real-time 
\texttt{/scan} and control messages 
(\texttt{/cmd\_vel}) and calculate
\texttt{/time\_to\_collision}. Central to \name, 
the \texttt{coordinator} node orchestrates the 
system, adjusting the CPU/GPU frequencies and ROS 
parameters in response to the data received from 
other nodes. Table~\ref{tab:robo-ems-parameters} 
presents the type and range for all the ROS 
parameters. The \texttt{coordinator} node updates 
\texttt{yolo\_wait} to manage YOLOv3's inference 
frequency. Similarly, it updates 
\texttt{controller\_frequency} and \texttt{max\_particles} 
to manage \texttt{amcl} and \texttt{dwa\_local\_planner}.


%% file: contents/6_Evaluation.tex
\label{sec:evaluation}

A combination of real-world AMR deployment and 
Gazebo simulation environments are utilized to evaluate
\name\ using metrics like power prediction accuracy, 
power savings, and navigation efficiency. 
Our key findings are the following:
\begin{itemize}[noitemsep,topsep=0pt,parsep=0pt,partopsep=0pt]
    \item \name\ is highly accuate 
in capturing and predicting the e2e power 
consumption of AMRs. Therefore, it can operate AMRs
within an RPM range that not only ensures reliable motor 
power prediction but also optimizes power-efficiency (\S\ref{subsec:power-prediction}).
    \item \name\ makes substantial 
improvements of power-efficiency for AMRs by intelligently 
adapting frequency settings and navigation configuration 
according to environmental conditions (\S\ref{subsec:power-reduction}).
    \item The design of \name\ demonstrates 
efficient navigation status monitoring, providing not 
only sufficient and safe navigation but also maintaining 
operation-efficiency. More importantly, it enhances the 
safety of AMRs, demonstrating \name\
as an efficient navigation tool (\S\ref{subsec:navigation-efficiency}).
\end{itemize}

\subsection{Experiment Setup}
\label{subsec:exp-setup}

We conduct fine-grained profiling using real robots and 
a realistic environment simulator. Profiling real AMRs 
provides a detailed power-consumption breakdown in 
different testing cases, while profiling with 
the simulator enables the evaluation of
power-efficiency of the software pipeline. 
Utilizing the Turtlebot model within the Gazebo 
environment simulator, we simulate a scenario where the 
Turtlebot Waffle navigates autonomously with the ROS 
Navigation Stack~\cite{ros-navigation} between designated 
points, incorporating an object detector based on YOLOv3 
to enhance navigation accuracy~\cite{redmon2018yolov3}.

\vspace{1mm}
\noindent\textbf{Metrics.} Our evaluation focuses on 
component-wise latency (including AMCL, Costmap2D, and 
DWA planner), power consumption across different modules 
(CPU, GPU, etc.), and localization accuracy, measuring both 
position and orientation errors defined in 
Eqs.~(\ref{equ:position-error}) and
(\ref{equ:orientation-error}). 
$(x_{\text{est}}, y_{\text{est}})$ and 
$\text{yaw}_{\text{est}}$ denote the robot's 
estimated position and orientation by SLAM while 
$(x_{\text{gt}}, y_{\text{gt}})$ and 
$\text{yaw}_{\text{gt}}$ are its real pose.
\begin{align}
    \label{equ:position-error}
    \text{Position Error} = \sqrt{(x_{\text{est}} - x_{\text{gt}})^2 + (y_{\text{est}} - y_{\text{gt}})^2}
\end{align}
\begin{align}
    \label{equ:orientation-error}
    \text{Orientation Error} = |\text{yaw}_{\text{est}} - \text{yaw}_{\text{gt}}|.
\end{align}

A software-based power monitor called \textsf{tegrastats} 
is leveraged for real-time monitoring of power consumption 
by different components on the Jetson AGX board. 
The Jetson AGX Xavier module has two 3-channel INA3221 
onboard power monitors for measuring the module's total 
power consumption. The measurement error of the power 
monitors is less than 5\%. For CPU, GPU, SOC, etc., 
\textsf{tegrastats} provides 
current/average power consumption in milliwatts.

\vspace{1mm}
\noindent\textbf{Testing Cases.} To comparatively evaluate \name, 
we establish four testing cases that incorporate common cyber 
and physical energy-saving solutions.

\vspace{1mm}
\noindent\textbf{Case \#1 (\texttt{SP}): Shortest Path with 
Default DVFS.} For the physical subsystem, both baselines' path 
planners use the collision-free shortest path, with all other 
parameters set to their default values as specified in ROS 
Navigation~\cite{turtlebot-github}. For the cyber subsystem, 
\texttt{SP} uses the default DVFS governors to adjust CPU and 
GPU frequencies, employing \texttt{scheutil} for CPUs and 
\texttt{nvhost\_podgov} for GPUs.

\vspace{1mm}
\noindent\textbf{Case \#2 (\texttt{SP+uDVFS}): Shortest Path with 
Utilization-Aware DVFS.} The physical subsystem is identical to 
that of Baseline~\#1. For the cyber subsystem, we design a 
utilization-aware DVFS governor similar to ~\cite{lin2023workload}. Specifically, all available CPU 
and GPU frequencies are sorted into a list. When the utilization 
of the CPU or GPU exceeds 80\%, the frequency is increased by 
one step in the list; conversely, if the utilization falls 
below 30\%, the frequency is decreased by one step.

\vspace{1mm}
\noindent\textbf{\texttt{pNav+uDVFS}}: Override the frequency 
configruation in \texttt{pNav} with the Utilization-Aware DVFS.

We evaluate \name\ and the other cases by running the robots for 
five loops through predefined waypoints.



\vspace{-2mm}
\subsection{Prediction of Power Consumption}
\label{subsec:power-prediction}

Our power prediction model forecasts the motor's power 
consumption through an MLP model while the power
consumption of the cyber subsystem is predicted
using a linear model. Below we first present the  
performance of the MLP model for motor's power prediction
and then the performance of predicting the AMR's total 
power consumption.

\subsubsection{Motor Power Prediction}

\begin{figure}[!htp]
	\centering
	\includegraphics[width=\columnwidth]{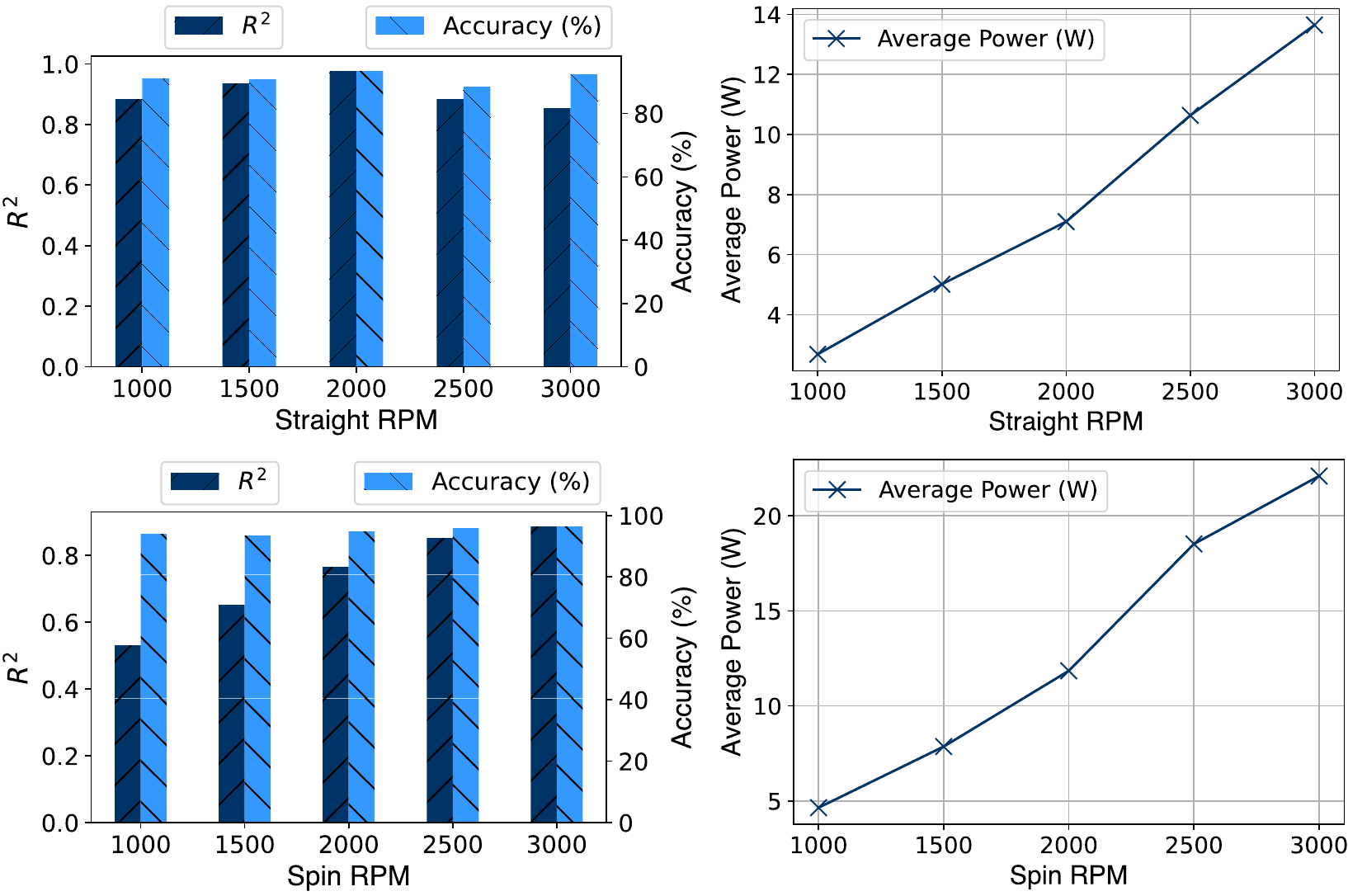}
\caption{The $R^2$ value and $accuracy$ of the  motor 
power prediction model and the predicted average power 
under different RPMs.}
	\label{fig:motor-power-prediction}
\end{figure}

As illustrated in Fig.~\ref{fig:motor-power-prediction}, 
our motor power prediction model is shown to yield
good accuracy and $R^2$ values across different 
RPMs. Specifically, the model's accuracy exceeded 90\% in 
both straight and spinning movements. The $R^2$ 
values were consistently above 0.8 for all RPMs during 
straight movement and exceeded 0.8 for spinning at RPMs 
of 2000 or higher. Furthermore, one can see a clear linear 
pattern in the predicted average power with respect
to RPM, indicating a direct and predictable relationship 
between these variables. This linear relationship was 
more pronounced in the range from 1000 to 3000 RPM, 
corroborating the effectiveness of our MLP model 
configuration in capturing and predicting the dynamics 
of the motor's power consumption.

\begin{table}[!ht]
\caption{Average cyber subsystem's power consumption 
and prediction accuracy at different GPU frequencies.}
\label{table:gpu_power_accuracy}
\centering
\resizebox{.85\columnwidth}{!}{%
\begin{tabular}{|c|c|c|}
\hline
\textbf{GPU (MHz)} & \textbf{Predicted Power (W)} & \textbf{Accuracy (\%)} \\
\hline
420 & 16.91 & \textbf{99.45} \\
624 & 18.92 & \textbf{99.31} \\
828 & 21.53 & \textbf{99.21} \\
1032 & 24.89 & \textbf{98.80} \\
1198 & 28.45 & \textbf{98.71} \\
1377 & 33.38 & \textbf{98.47} \\
\hline
\end{tabular}
}
\end{table}

\subsubsection{E2E AMR Power Prediction}

For power-sumption prediction for cyber/embedded systems, 
we initially collected offline power traces to establish 
a model, as in Eq.~(\ref{equ:embedded-power}). 
From our empirical study, we observed a linear 
relationship between the power consumption and 
operating frequencies of CPU and GPU. 
To validate this model, we configured a Jetson board at 
various CPU and GPU frequencies and ran our software 
pipeline, measuring the actual power consumption. 
The average of both predicted and measured power 
consumption was then calculated. The results, as shown 
in Table~\ref{table:gpu_power_accuracy}, show that 
our power prediction model achieves over 98\% accuracy 
across all GPU frequencies.

\begin{table}[!ht]
\caption{E2E power consumption, error, and accuracy for 
different spin RPMs and straight motion speeds.}
\label{tab:power_stats}
\centering
\resizebox{\columnwidth}{!}{%
\begin{tabular}{|c|c|c|c|}
\hline
\textbf{Spin RPM} & \textbf{e2e AMR Power (W)} & \textbf{Error (W)} & \textbf{Accuracy (\%)} \\
\hline
1000 & 51.96 & 1.63 & \textbf{96.86} \\
1500 & 64.82 & 2.59 & \textbf{96.01} \\
2000 & 80.76 & 3.08 & \textbf{96.18} \\
\hline
\textbf{Straight RPM} & \textbf{e2e AMR Power (W)} & \textbf{Error (W)} & \textbf{Accuracy (\%)} \\
\hline
1000 & 44.11 & 1.48 & \textbf{96.64} \\
1500 & 53.44 & 2.37 & \textbf{95.56} \\
2000 & 61.77 & 2.37 & \textbf{96.15} \\
\hline
\end{tabular}
}
\end{table}

Using the power consumption predictions for both 
the motors and the cyber subsystem, we developed a 
comprehensive e2e power consumption prediction model. 
Table~\ref{tab:power_stats} provides the e2e power 
consumption numbers, along with the error and accuracy 
numbers, for both spinning and straight-line movements 
at various RPMs. These results consistently show 
$>95$\% accuracies across all test cases, corroborating 
the effectiveness of \name's power prediction module.


\vspace{-2mm}
\subsection{Power Consumption}
\label{subsec:power-reduction}

To evaluate the effectiveness of \name\ in minimizing power 
consumption, we conducted tests with robots operating under 
both the baseline configurations and \name\ settings, 
executing identical sets of five goal loops. 
The power consumption of the cyber subsystem was recorded, 
and the results are summarized in 
Table~\ref{tab:power-dissipation-breakdown}. 
This data reveals that \name\ significantly reduces power 
consumption across all components by dynamically adjusting 
the CPU and GPU frequencies. Notably, the total power 
consumption was reduced from 60.41W with the testing case 
\texttt{SP} to 37.38W with \name, marking a substantial (38.1\%) 
reduction of power consumption. Even comparing with the testing case
\texttt{SP+uDVFS}, \texttt{pNav} still reduces over 31\% of power.

\begin{table}[!ht]
\caption{Power Consumption Comparison.}
\label{table:power-comparison-reduction}
\centering
\resizebox{\columnwidth}{!}{%
\begin{tabular}{|c|c|c|c|c|c|c|}
\hline
\textbf{Power (W)} & \textbf{\texttt{SP}} & \textbf{\makecell{\texttt{SP}\\\texttt{+uDVFS}}} & \textbf{\makecell{\texttt{pNav}\\\texttt{+uDVFS}}} & \textbf{\name} & \textbf{\makecell{Reduction\\over \texttt{SP} (\%)}} & \textbf{\makecell{Reduction over\\ \texttt{SP+uDVFS} (\%)}} \\
\hline
GPU    & 15.19 & 10.3  & 5.8  & 2.22 & 85.41\% & 78.45\% \\
CPU    & 8.92  & 8.6   & 9.2  & 7.72 & 13.41\% & 10.23\% \\
SOC    & 5.24  & 4.7   & 4.2  & 3.85 & 26.58\% & 18.09\% \\
VDDRQ  & 2.87  & 2.6   & 2    & 1.66 & 42.21\% & 36.15\% \\
SYS5V  & 4.89  & 4.7   & 4.4  & 4.24 & 13.36\% & 9.79\% \\ \hline
Motor  & 23.3  & 23.3  & 17.7 & 17.7 & 22.5\% & 24.03\% \\ 
\hline
\rowcolor[gray]{.85} 
\textbf{Total} & \textbf{60.41} & \textbf{54.2} & \textbf{43.3} & \textbf{37.38} & \textbf{38.1\%} & \textbf{31.03\%} \\
\hline
\end{tabular}
}
\end{table}

\subsection{Navigation Efficiency}
\label{subsec:navigation-efficiency}

In addition to the significant reduction of
power consumption, \name\ is expected to achieve good 
navigation performance since it monitors the environment 
and dynamically configures its execution in real time. 
So, we evaluate the navigation efficiency of \name\ in 
task latency, loop finish time, and navigation safety. 

\subsubsection{Task Latency}

\begin{figure}[!htp]
	\centering
	\includegraphics[width=\columnwidth]{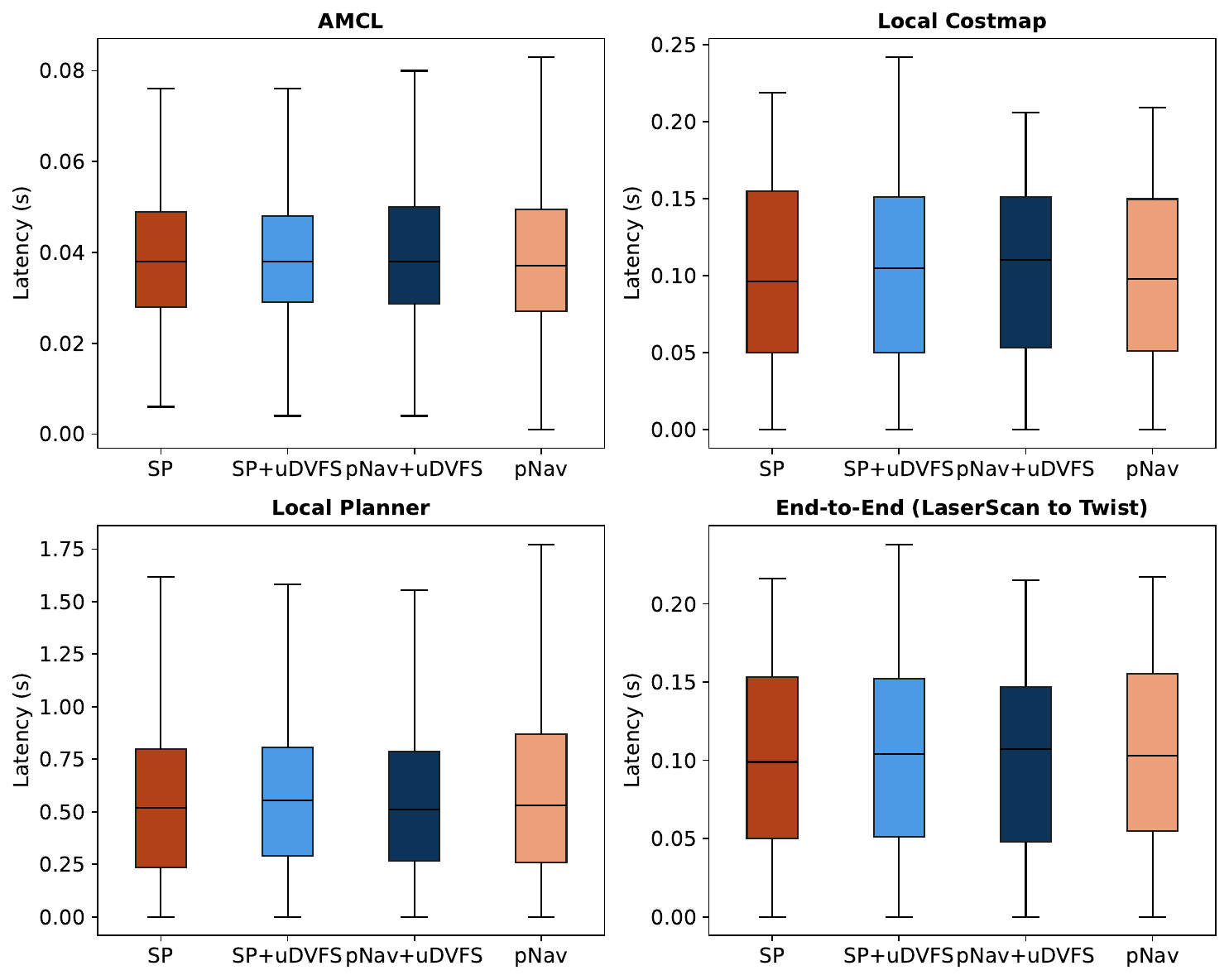}
	\caption{Boxplot of navigation task latency 
    for the baseline and \name.}
	\label{fig:latency-boxplot}
\end{figure}

Latency is crucial for the efficiency and safety of 
navigation. Fig.~\ref{fig:latency-boxplot} compares the 
latency boxplot for the AMCL, local costmap, and local 
planner components between the baseline and \name. 
These components are key contributors to the e2e navigation 
timeline of AMRs. Our analysis reveals that \name\ maintains 
a similar average latency and distribution to the baseline.

Further insights into the average latency and its changes 
are provided in Table~\ref{tab:latency_table}. 
\name\ alters the average e2e latency by a maximum of 1.75\%, 
equivalent to approximately 2ms. This marginal change is 
negligible in the AMR's safe control.

\begin{table}[!ht]
\caption{Average latency (in milliseconds) for each component across different testing cases.}
\label{tab:latency_table}
\centering
\resizebox{\columnwidth}{!}{%
\begin{tabular}{|c|c|c|c|c|c|}
\hline
\textbf{Latency (ms)} & \textbf{SP} & \textbf{\makecell{SP\\+uDVFS}} & \textbf{\makecell{pNav\\+uDVFS}} & \textbf{pNav} & \textbf{\makecell{\% Change\\over \texttt{SP}}} \\
\hline
AMCL & 40.608 & 39.154 & 40.638 & 38.470 & -5.16\% \\
Local Costmap & 101.043 & 101.094 & 104.016 & 99.300 & -1.72\% \\
Local Planner & 565.647 & 604.053 & 573.911 & 628.375 & +11.03\% \\
\hline
\rowcolor[gray]{.85} \textbf{End-to-End} & \textbf{101.420} & \textbf{102.134} & \textbf{100.746} & \textbf{103.176} & \textbf{+1.75\%} \\
\hline
\end{tabular}
}
\end{table}

\subsubsection{Loop Finish Time}

The completion time for a loop of goals is critical 
metric in assessing the operation-efficiency of AMRs. 
To this end, we collected finish times for 20 goal points 
under both the baseline and \name\ configurations. 
As shown in Table~\ref{tab:finish_times}, there are 
noticeable differences in the average, minimum, and 
maximum finish times between the two systems. 
While the baseline exhibits a higher maximum finish time, 
\name\ provides a lower minimum finish time. 

\begin{table}[!ht]
\centering
\caption{Comparison of Finish Times.}
\label{tab:finish_times}
\resizebox{.8\columnwidth}{!}{%
\begin{tabular}{|c|c|c|c|}
\hline
\textbf{Finish Time (s)} & \textbf{Average} & \textbf{Min} & \textbf{Max} \\
\hline
\textbf{\texttt{SP}} & 22.64 & 18.41 & 25.34 \\
\hline
\textbf{\texttt{SP+uDVFS}} & 21.92 & 18.09 & 33.54 \\
\hline
\textbf{\texttt{pNav+uDVFS}} & 22.80 & 17.98 & 25.92 \\
\hline
\rowcolor[gray]{.85} \name\ & 20.50 & 16.47 & 24.14 \\
\hline
\end{tabular}
}
\end{table}

Overall, \name\ reduces the average finish time by 2.14 seconds 
(9.4\%) compared to the \texttt{SP} case. This improvement 
demonstrates \texttt{pNav}'s navigation efficiency. The increased 
performance is largely due to \texttt{pNav}'s power-efficient 
planner, which optimizes speed, heading, and distance in a 
comprehensive manner.

\subsubsection{Navigation Safety}

\begin{figure}[!htp]
	\centering
	\includegraphics[width=\columnwidth]{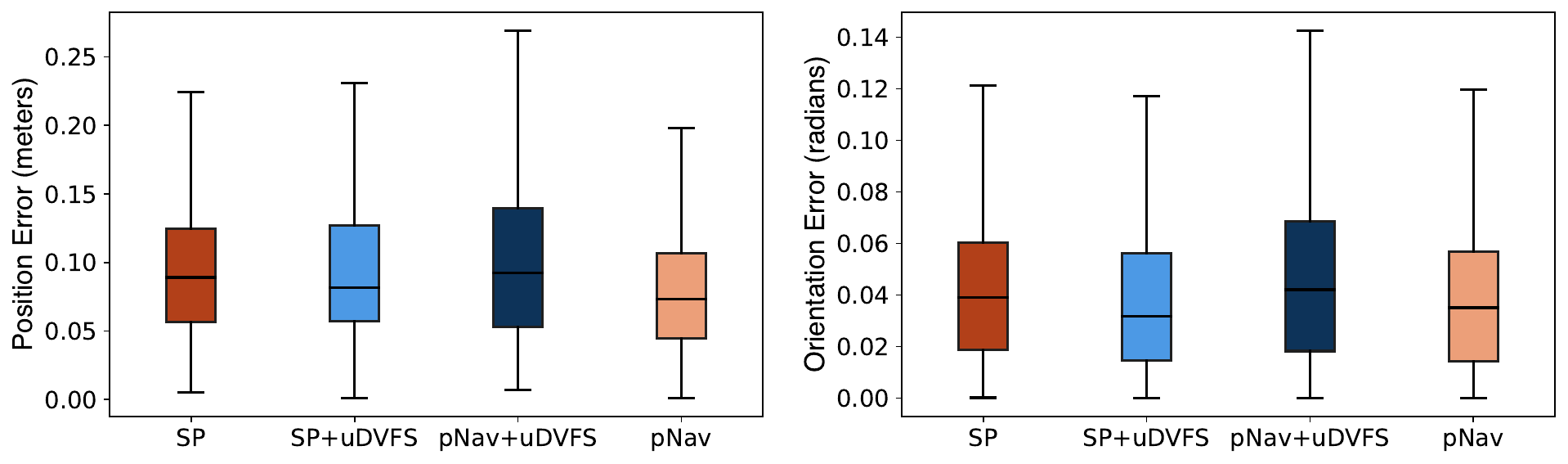}
\caption{Boxplot of AMCL accuracy for the baseline and \name.}
	\label{fig:accuracy-compare-boxplot}
\end{figure}

Our evaluation of navigation safety focuses on two 
key metrics: localization accuracy and TTC. These metrics 
are crucial for ensuring the correct localization and 
adequate reaction time of an AMR. 
Fig.~\ref{fig:accuracy-compare-boxplot} presents a boxplot 
comparison of AMCL's positional and orientation errors 
between all testing cases and \name. The results show a similar 
average and distribution of errors for both systems. 

Table~\ref{tab:error_comparison} provides the average 
positional and orientation errors, along with the percentage 
changes. \name\ demonstrates a 18--38\% reduction in 
error rates compared to the testing case \texttt{SP}. 
Compared to testing case \texttt{SP+uDVFS}, \texttt{pNav} 
also reduces 13--50\% position and orientation errors.


\begin{table}[!ht]
\centering
\caption{Position and Orientation Errors.}
\label{tab:error_comparison}
\resizebox{.8\columnwidth}{!}{%
\begin{tabular}{|c|c|c|}
\hline
\textbf{Cases} & \textbf{\makecell{Position Error\\(meters)}} & \textbf{\makecell{Orientation Error\\(radians)}} \\ \hline

\textbf{\texttt{SP}}            & 0.100    & 0.063   \\ \hline
\textbf{\texttt{SP+uDVFS}}      & 0.095    & 0.078   \\ \hline
\textbf{\texttt{pNav+uDVFS}}    & 0.107    & 0.103   \\ \hline

\rowcolor[gray]{0.85} 
\texttt{pNav}           & 0.082    & 0.039   \\ \hline

\rowcolor[gray]{0.85} 
\textbf{\% Change over \texttt{SP}} & \textbf{-18.09} & \textbf{-38.25} \\ \hline
\rowcolor[gray]{0.85} 
\textbf{\begin{tabular}[c]{@{}c@{}}\% Change over\\ \texttt{SP+uDVFS}\end{tabular}} & \textbf{-13.84} & \textbf{-50.12} \\ \hline
\end{tabular}%
}
\end{table}

TTC is another crucial safety metric. 
A longer TTC provides the AMR with more time to react and 
avoid potential collision. We deployed an additional node 
to record TTC data at 5Hz for both the baseline 
and \name\ configurations. Table~\ref{tab:time_to_collision} 
summarizes the average, minimum, and maximum TTC. 
The data reveals that \name\ reduces the average 
TTC by 0.04 second and the minimum TTC by 1.13 second than 
\texttt{SP}, which is within an acceptable limit for 
AMR operation. Moreover, \name\ extends the maximum time by 0.44 
second than \texttt{SP}. This is achieved mainly by integrating power 
prediction into path selection (Eq.~(\ref{equ:dwa-with-power})), 
which reduces the bias for higher speeds.


\begin{table}[!ht]
\centering
\caption{Time to Collision.}
\label{tab:time_to_collision}
\resizebox{.8\columnwidth}{!}{%
\begin{tabular}{|c|c|c|c|}
\hline
\rowcolor[gray]{0.85} 
\textbf{Cases} & \textbf{Average (s)} & \textbf{Min (s)} & \textbf{Max (s)} \\ \hline

\texttt{SP}       & 6.25 & 1.84 & 12.98 \\ \hline
\texttt{SP+uDVFS}      & 6.44 & 2.61 & 11.60 \\ \hline
\texttt{pNav+uDVFS}    & 6.22 & 1.42 & 13.24 \\ \hline

\rowcolor[gray]{0.85} 
\texttt{pNav}           & 6.21 & 0.71 & 13.42 \\ \hline

\rowcolor[gray]{0.85} 
\textbf{\% Change over \texttt{SP}} & \textbf{-0.04} & \textbf{-1.13} & \textbf{+0.44} \\ \hline
\rowcolor[gray]{0.85} 
\textbf{\begin{tabular}[c]{@{}c@{}}\% Change over\\ \texttt{SP+uDVFS}\end{tabular}} & \textbf{-0.23} & \textbf{-1.9} & \textbf{+1.82} \\ \hline

\end{tabular}%
}
\end{table}


%% file: contents/8_Related_Works.tex
\label{sec:related-work}

Various strategies have been proposed to extend the battery 
life of AMRs. These can be broadly categorized into two 
main areas: energy-efficient motion planning and optimizing 
the power-efficiency of embedded systems.

\vspace{-2mm}
\subsection{Energy-Efficient Motion Planning}

Mechanical components, particularly motors, typically 
account for about 50\% of an AMR's total power consumption. 
So, considerable research has focused on devising the most 
energy-efficient paths for robots, either for reaching 
specific destinations or for area coverage 
\cite{path_planning1, path_planning3, path_planning4, 
movingsensor, movingsensor2, movingsensor3}. These studies 
generally model the correlation between motor speed and 
power usage, incorporating this model into the robot’s 
navigation algorithms \cite{mei2004energy, 
liu2013minimizing}. However, a common limitation of these 
studies are their predominant focus on the robots' 
mechanical power consumption, often overlooking or 
oversimplifying the power demands of computing systems. 
They become increasingly inaccurate with the growing use of 
GPUs in embedded systems, which significantly increases 
power consumption \cite{jetson-comparison}.

\vspace{-3mm}
\subsection{Power-Efficient Cyber Subsystems}

Cyber/embedded systems are pivotal in AMRs as they host all 
the navigation software. In recent years, numerous studies 
attempted to optimize the power consumption of embedded 
systems, particularly using Dynamic Voltage and Frequency Scaling (DVFS) 
\cite{chen2015smartphone, lin2023workload, choi2019optimizing, bateni2020neuos, choi2019graphics}. 
Some have leveraged application (e.g., games, 
browsers) profiling combined with DVFS to 
reduce power consumption in smartphones \cite{chen2015smartphone, choi2019optimizing}. 
Others have concentrated on optimizing the power consumption 
of Deep Neural Networks (DNNs) running on embedded systems 
equipped with GPUs \cite{bateni2020neuos}. However, these 
solutions often fall short for AMRs due to their 
distinct workload characteristics. E2M optimizes the power 
consumption of embedded systems in AMRs, striking a balance 
between application performance and minimal power usage of 
computing resources \cite{liu2019e2m}. However, it largely 
overlooked the mechanical aspects of AMRs. 
Another study proposes an e2e energy model for AMRs 
\cite{liu2023open}, but it focues on average energy consumption
and its path planning is overly simplistic, lacking a 
fine-grained power management.

\name\ provides an orthogonal (hence an alternative) approach to SOTA 
works on isolated cyber or physical subsystems by co-optimizing both, 
to achieve better power-efficiency. It extends beyond its initial scope 
to serve a variety of mobile cyber-physical systems, enhancing tasks from 
lawn-mowing to managing automated guided vehicles, and even aiding 
in search \& rescue, and battlefield operations.

%% file: contents/9_Conclusion.tex
\label{sec:conclusion}

We have presented a novel power management system for 
Autonomous Mobile Robots (AMRs), called \name. 
It offers a holistic approach that 
accounts for the power demands by both the physical 
and cyber parts of AMRs. 
Using detailed power profiling of key components and 
addressing the challenges in balancing 
e2e power consumption, modeling navigation locality, 
and coordination of cyber-physical subsystems, \name\ not only 
enhances the AMR's power-efficiency but also 
ensures the safe and efficient operation of AMRs. 
The implementation of \name\ within the ROS Navigation 
Stack is shown to reduce the AMR's power consumption 
by 38.1\% and achieved over 96\% accuracy in 
power-consumption prediction. Moreover, \name's ability to 
extend the time-to-collision (TTC) and maintain navigation 
performance with minimal latency and accuracy impact 
corroborates its effectiveness in monitoring 
and controlling modern AMRs.

%% file: main.bib
@ARTICLE{movingsensor3, 
author={Y. {Wang} and W. {Peng} and Y. {Tseng}}, 
journal={IEEE Transactions on Parallel and Distributed Systems}, 
title={Energy-Balanced Dispatch of Mobile Sensors in a Hybrid Wireless Sensor Network}, 
year={2010}, 
volume={21}, 
number={12}, 
pages={1836-1850}, 
doi={10.1109/TPDS.2010.56}, 
ISSN={1045-9219}, 
month={Dec},}

@article{movingsensor2,
 author = {Anagnostopoulos, Christos and Hadjiefthymiades, Stathes and Kolomvatsos, Kostas},
 title = {Accurate, Dynamic, and Distributed Localization of Phenomena for Mobile Sensor Networks},
 journal = {ACM Trans. Sen. Netw.},
 issue_date = {May 2016},
 volume = {12},
 number = {2},
 month = apr,
 year = {2016},
 issn = {1550-4859},
 pages = {9:1--9:59},
 articleno = {9},
 numpages = {59},
 url = {http://doi.acm.org/10.1145/2882966},
 doi = {10.1145/2882966},
 acmid = {2882966},
 publisher = {ACM},
 address = {New York, NY, USA},
}

@INPROCEEDINGS{movingsensor, 
author={S. {Yu} and C. S. G. {Lee}}, 
booktitle={2011 IEEE International Conference on Robotics and Automation}, 
title={Lifetime maximization in mobile sensor networks with energy harvesting}, 
year={2011}, 
volume={}, 
number={}, 
pages={5911-5916}, 
doi={10.1109/ICRA.2011.5979588}, 
ISSN={1050-4729}, 
month={May},}

@INPROCEEDINGS{recog, 
author={D. {Avola} and G. L. {Foresti} and L. {Cinque} and C. {Massaroni} and G. {Vitale} and L. {Lombardi}}, 
booktitle={2016 IEEE 14th International Conference on Industrial Informatics (INDIN)}, 
title={A multipurpose autonomous robot for target recognition in unknown environments}, 
year={2016}, 
volume={}, 
number={}, 
pages={766-771}, 
doi={10.1109/INDIN.2016.7819262}, 
ISSN={2378-363X}, 
month={July},}

@article{path_planning4,
title = "Energy Efficient Dynamic Window Approach for Local Path Planning in Mobile Service Robotics**This work was conducted at the {University of Auckland, Auckland, New Zealand}",
journal = "IFAC-PapersOnLine",
volume = "49",
number = "15",
pages = "32 - 37",
year = "2016",
note = "9th IFAC Symposium on Intelligent Autonomous Vehicles IAV 2016",
issn = "2405-8963",
doi = "https://doi.org/10.1016/j.ifacol.2016.07.610",
url = "http://www.sciencedirect.com/science/article/pii/S2405896316308813",
author = "Christian Henkel and Alexander Bubeck and Weiliang Xu",
}

@Article{path_planning3,
author="Broderick, John A.
and Tilbury, Dawn M.
and Atkins, Ella M.",
title="Optimal coverage trajectories for a {UGV} with tradeoffs for energy and time",
journal="Autonomous Robots",
year="2014",
month="Mar",
day="01",
volume="36",
number="3",
pages="257--271",
issn="1573-7527",
doi="10.1007/s10514-013-9348-x",
url="https://doi.org/10.1007/s10514-013-9348-x"
}

@INPROCEEDINGS{path_planning1, 
author={S. {Dogru} and L. {Marques}}, 
booktitle={2015 IEEE International Conference on Autonomous Robot Systems and Competitions}, 
title={Energy Efficient Coverage Path Planning for Autonomous Mobile Robots on {3D} Terrain}, 
year={2015}, 
volume={}, 
number={}, 
pages={118-123}, 
doi={10.1109/ICARSC.2015.23}, 
ISSN={}, 
month={April},}

@inproceedings{AMR_DC,
 author = {Lenchner, Jonathan and Isci, Canturk and Kephart, Jeffrey O. and Mansley, Christopher and Connell, Jonathan and McIntosh, Suzanne},
 title = {Towards Data Center Self-diagnosis Using a Mobile Robot},
 booktitle = {Proceedings of the 8th ACM International Conference on Autonomic Computing},
 series = {ICAC '11},
 year = {2011},
 isbn = {978-1-4503-0607-2},
 location = {Karlsruhe, Germany},
 pages = {81--90},
 numpages = {10},
 url = {http://doi.acm.org/10.1145/1998582.1998597},
 doi = {10.1145/1998582.1998597},
 acmid = {1998597},
 publisher = {ACM},
 address = {New York, NY, USA},
}

@INPROCEEDINGS{AMR_farm, 
author={H. {Durmuş} and E. O. {Güneş} and M. {Kırcı} and B. B. {Üstündağ}}, 
booktitle={2015 Fourth International Conference on Agro-Geoinformatics (Agro-geoinformatics)}, 
title={The design of general purpose autonomous agricultural mobile-robot: “{AGROBOT}”},
year={2015}, 
volume={}, 
number={}, 
pages={49-53}, 
doi={10.1109/Agro-Geoinformatics.2015.7248088}, 
ISSN={}, 
month={July},}

@misc{jetson-comparison,
  title = {{Jetson Modules}},
  note = {\url{https://developer.nvidia.com/embedded/jetson-modules}},
  year={2021}
}

@misc{ros-dwa-planner,
  title = {{dwa\_local\_planner}},
  note = {\url{http://wiki.ros.org/dwa\_local\_planner}},
  year={2024}
}

@misc{ros-amcl,
  title = {{amcl}},
  note = {\url{http://wiki.ros.org/amcl}},
  year={2024}
}

@misc{ros-navigation,
  title = {{ROS Navigation Stack}},
  note = {\url{https://github.com/ros-planning/navigation}},
  year={2024}
}

@misc{turtlebot-github,
  title = {{turtlebot3}},
  note = {\url{https://github.com/ROBOTIS-GIT/turtlebot3}},
  year={2024}
}

@article{redmon2018yolov3,
  title={{YOLOv3:} An incremental improvement},
  author={Redmon, Joseph and Farhadi, Ali},
  journal={arXiv preprint arXiv:1804.02767},
  year={2018}
}

@misc{allegro_microsystems_acs712_2021,
	title = {{ACS712}: {Hall}-{Effect}-{Based} {Linear} {Current} {Sensor} {IC}},
	url = {https://www.allegromicro.com/en/products/sense/current-sensor-ics/zero-to-fifty-amp-integrated-conductor-sensor-ics/acs712},
	language = {English},
	urldate = {2021-12-13},
	journal = {ACS712: Fully Integrated, Hall-Effect-Based Linear Current Sensor IC with 2.1 kVRMS Voltage Isolation and a Low-Resistance Current Conductor},
	author = {{Allegro MicroSystems}},
	month = dec,
	year = {2021},
}

@misc{texas_instruments_ina1x9_2021,
	title = {{INA1x9} {High}-{Side} {Measurement} {Current} {Shunt} {Monitor}},
	url = {https://www.ti.com/lit/ds/symlink/ina169.pdf},
	language = {English},
	journal = {ti.com},
	author = {{Texas Instruments}},
	month = dec,
	year = {2021},
}

@inproceedings{liu2019e2m,
  title={E2M: an energy-efficient middleware for computer vision applications on autonomous mobile robots},
  author={Liu, Liangkai and Chen, Jiamin and Brocanelli, Marco and Shi, Weisong},
  booktitle={Proceedings of the 4th ACM/IEEE Symposium on Edge Computing},
  pages={59--73},
  year={2019}
}

@inproceedings{mei2004energy,
  title={Energy-efficient motion planning for mobile robots},
  author={Mei, Yongguo and Lu, Yung-Hsiang and Hu, Y Charlie and Lee, CS George},
  booktitle={IEEE International Conference on Robotics and Automation, 2004. Proceedings. ICRA'04. 2004},
  volume={5},
  pages={4344--4349},
  year={2004},
  organization={IEEE}
}

@article{fox1997dynamic,
  title={The dynamic window approach to collision avoidance},
  author={Fox, Dieter and Burgard, Wolfram and Thrun, Sebastian},
  journal={IEEE Robotics \& Automation Magazine},
  volume={4},
  number={1},
  pages={23--33},
  year={1997},
  publisher={IEEE}
}

@inproceedings{liu2023open,
  title={An Open Approach to Energy-Efficient Autonomous Mobile Robots},
  author={Liu, Liangkai and Zhong, Ren and Willcock, Aaron and Fisher, Nathan and Shi, Weisong},
  booktitle={2023 IEEE International Conference on Robotics and Automation (ICRA)},
  pages={11569--11575},
  year={2023},
  organization={IEEE}
}

@article{liu2013minimizing,
  title={Minimizing energy consumption of wheeled mobile robots via optimal motion planning},
  author={Liu, Shuang and Sun, Dong},
  journal={IEEE/ASME Transactions on Mechatronics},
  volume={19},
  number={2},
  pages={401--411},
  year={2013},
  publisher={IEEE}
}

@inproceedings{chen2015smartphone,
  title={Smartphone background activities in the wild: Origin, energy drain, and optimization},
  author={Chen, Xiaomeng and Jindal, Abhilash and Ding, Ning and Hu, Yu Charlie and Gupta, Maruti and Vannithamby, Rath},
  booktitle={Proceedings of the 21st Annual International Conference on Mobile Computing and Networking},
  pages={40--52},
  year={2015}
}

@inproceedings{lin2023workload,
  title={A workload-aware dvfs robust to concurrent tasks for mobile devices},
  author={Lin, Chengdong and Wang, Kun and Li, Zhenjiang and Pu, Yu},
  booktitle={Proceedings of the 29th Annual International Conference on Mobile Computing and Networking},
  pages={1--16},
  year={2023}
}

@inproceedings{choi2019optimizing,
  title={Optimizing energy efficiency of browsers in energy-aware scheduling-enabled mobile devices},
  author={Choi, Yonghun and Park, Seonghoon and Cha, Hojung},
  booktitle={The 25th Annual International Conference on Mobile Computing and Networking},
  pages={1--16},
  year={2019}
}

@inproceedings{bateni2020neuos,
  title={$NeuOS$: A Latency-Predictable Multi-Dimensional Optimization Framework for DNN-driven Autonomous Systems},
  author={Bateni, Soroush and Liu, Cong},
  booktitle={2020 USENIX Annual Technical Conference (USENIX ATC 20)},
  pages={371--385},
  year={2020}
}

@inproceedings{choi2019graphics,
  title={Graphics-aware power governing for mobile devices},
  author={Choi, Yonghun and Park, Seonghoon and Cha, Hojung},
  booktitle={Proceedings of the 17th annual international conference on mobile systems, applications, and services},
  pages={469--481},
  year={2019}
}

@book{siegwart2011introduction,
  title={Introduction to autonomous mobile robots},
  author={Siegwart, Roland and Nourbakhsh, Illah Reza and Scaramuzza, Davide},
  year={2011},
  publisher={MIT press}
}

@inproceedings{lee2008cyber,
  title={Cyber physical systems: Design challenges},
  author={Lee, Edward A},
  booktitle={2008 11th IEEE international symposium on object and component-oriented real-time distributed computing (ISORC)},
  pages={363--369},
  year={2008},
  organization={IEEE}
}

@article{henkel2016energy,
  title={Energy efficient dynamic window approach for local path planning in mobile service robotics},
  author={Henkel, Christian and Bubeck, Alexander and Xu, Weiliang},
  journal={IFAC-papersonline},
  volume={49},
  number={15},
  pages={32--37},
  year={2016},
  publisher={Elsevier}
}

@inproceedings{brateman2006energy,
  title={Energy-effcient scheduling for autonomous mobile robots},
  author={Brateman, Jeff and Xian, Changjiu and Lu, Yung-Hsiang},
  booktitle={2006 IFIP international conference on very large scale integration},
  pages={361--366},
  year={2006},
  organization={IEEE}
}

@article{farooq2023power,
  title={Power solutions for autonomous mobile robots: A survey},
  author={Farooq, Muhammad Umar and Eizad, Amre and Bae, Hyun-Ki},
  journal={Robotics and Autonomous Systems},
  volume={159},
  pages={104285},
  year={2023},
  publisher={Elsevier}
}

@inproceedings{barili1995energy,
  title={Energy-saving motion control for an autonomous mobile robot},
  author={Barili, A and Ceresa, M and Parisi, C},
  booktitle={1995 Proceedings of the IEEE International Symposium on Industrial Electronics},
  volume={2},
  pages={674--676},
  year={1995},
  organization={IEEE}
}

@inproceedings{swanborn2020energy,
  title={Energy efficiency in robotics software: A systematic literature review},
  author={Swanborn, Stan and Malavolta, Ivano},
  booktitle={Proceedings of the 35th IEEE/ACM International Conference on Automated Software Engineering},
  pages={144--151},
  year={2020}
}

@inproceedings{mei2005case,
  title={A case study of mobile robot's energy consumption and conservation techniques},
  author={Mei, Yongguo and Lu, Yung-Hsiang and Hu, Y Charlie and Lee, CS George},
  booktitle={ICAR'05. Proceedings., 12th International Conference on Advanced Robotics, 2005.},
  pages={492--497},
  year={2005},
  organization={IEEE}
}

@inproceedings{gatesichapakorn2019ros,
  title={ROS based autonomous mobile robot navigation using 2D LiDAR and RGB-D camera},
  author={Gatesichapakorn, Sukkpranhachai and Takamatsu, Jun and Ruchanurucks, Miti},
  booktitle={2019 First international symposium on instrumentation, control, artificial intelligence, and robotics (ICA-SYMP)},
  pages={151--154},
  year={2019},
  organization={IEEE}
}

@article{pimentel2021evaluation,
  title={Evaluation of ROS navigation stack for social navigation in simulated environments},
  author={Pimentel, Fagner de Assis Moura and Aquino-Jr, Plinio Thomaz},
  journal={Journal of Intelligent \& Robotic Systems},
  volume={102},
  number={4},
  pages={87},
  year={2021},
  publisher={Springer}
}

@inproceedings{gaskell2024mbot,
  title={MBot: A modular ecosystem for scalable robotics education},
  author={Gaskell, Peter and Pavlasek, Jana and Gao, Tom and Narula, Abhishek and Lewis, Stanley and Jenkins, Odest Chadwicke},
  booktitle={2024 IEEE International Conference on Robotics and Automation (ICRA)},
  pages={18294--18300},
  year={2024},
  organization={IEEE}
}
